\pdfoutput=1

\documentclass[11pt]{article}
\usepackage{array}
\usepackage{censor}
\usepackage{paralist}
\usepackage{xcolor}
\usepackage{colortbl}
\usepackage{tikz}
\usepackage{pgfplots}
\usepackage{booktabs}
\usepackage{multirow}
\usepackage{placeins}
\usepackage{amsmath}
\usepackage{subcaption}
\usepackage{comment}
\usepackage{float}
\usepackage{mathtools}

\pgfplotsset{compat=newest}
\usepackage[final]{acl}

\usepackage{times}
\usepackage{latexsym}
 
\usepackage[T1]{fontenc}

\usepackage[utf8]{inputenc}

\usepackage{microtype}

\usepackage{inconsolata}

\usepackage{graphicx}

\title{Hope vs. Hate: Understanding  User Interactions with LGBTQ+ News Content in Mainstream US News Media through the Lens of Hope Speech}

\author{Jonathan Pofcher$^{\clubsuit}$ \hspace{0.2cm} Christopher M. Homan$^{\clubsuit}$ \hspace{0.3cm} \\\textbf{Randall L. Sell$^{\diamondsuit}$} \hspace{0.2cm} 
\hspace{0.2cm}  \textbf{Ashiqur R. KhudaBukhsh$^{\clubsuit}$}\thanks{\hspace{0.2cm} Ashiqur R. KhudaBukhsh is the corresponding author.} \\
  $^{\clubsuit}$Rochester Institute of Technology \\
  $^{\diamondsuit}$Drexel University \\
  \texttt{jep7453@rit.edu}, \texttt{cmhvcs@rit.edu},  \texttt{randy@drexel.edu}, \texttt{axkvse@rit.edu}  \\}

\begin{document}
\maketitle
\begin{abstract}
This paper makes three contributions. 
First, via a substantial corpus of 1,419,047 comments posted on 3,161 YouTube news videos of major US cable news outlets, we analyze how users engage with LGBTQ+ news content. Our analyses focus both on positive and negative content. In particular, we construct a \textit{hope speech} classifier that detects positive (\textit{hope speech}), negative, neutral, and irrelevant content. Second, in consultation with a public health expert specializing on LGBTQ+ health, we conduct an annotation study with a balanced and diverse political representation and release a dataset of 3,750 instances with crowd-sourced labels and detailed annotator demographic information. Finally, beyond providing a vital resource for the LGBTQ+ community, our annotation study and subsequent in-the-wild assessments reveal (1) strong association between rater political beliefs and how they rate content relevant to a marginalized community, (2) models trained on individual political beliefs exhibit considerable in-the-wild disagreement, and (3) zero-shot large language models (LLMs) align more with liberal raters.   


\textbf{\textcolor{red}{Trigger Warning: this paper contains offensive material that some may find upsetting.}}

\end{abstract}

\section{Introduction}

From crowdsourced storymapping projects sharing stories of love, loss, and a sense of belonging~\cite{kirby2021queering} to safe, anonymous spaces to seek resources~\cite{mcinroy2019lgbtq+} and dating platforms~\cite{blackwell2015seeing} -- the internet and modern technologies play a positive role in the health and well-being of the LGBTQ+ community in various ways. However, cyberbullying~\cite{abreu2018cyberbullying}, exposure to dehumanization through news media~\cite{Mendelsohn_2020}, and more recently,  homophobic biases in large language models (LLMs)~\cite{IJCAI2024RabbitHole,DBLP:conf/aaai/DuttaPK25} - are some of the modern technology perils the community grapples with. 

\textit{How do mainstream US cable news outlets cover the LGBTQ+ community and how do users interact with such news content?} Via a substantial corpus of more than 130 million YouTube comments on more than 300K YouTube news videos uploaded on three major US cable news outlets, this paper investigates how LGBTQ+ discussions are situated in mainstream US political discourse. While prior literature has analyzed LGBTQ+ discourse in print media through the lens of a dehumanization framework~\cite{Mendelsohn_2020}, such analyses consulted a single source of news (The New York Times) and did not consider reader responses to LGBTQ+ news. In this work, we consider three major US cable news networks representing diverse political views (Fox News, CNN, and MSNBC) with a key goal to understand how users interact and engage with LGBTQ+ news content. In doing so, we not only focus on identifying negative discussions about the LGBTQ+ community, one of our key contributions is operationalizing the detection of \textit{hope speech} championing the disadvantaged minority in the broader spirit of counter-speech literature~\cite{benesch2014countering,mathew2019thou,palakodety2019hope,chakravarthi-2020-hopeedi,DBLP:conf/naacl/HenglePSBA024}.

\begin{table*}[t]
\centering
\small
\begin{tabular}{|p{0.85\textwidth}|p{0.1\textwidth}|}
\hline
\textbf{YouTube Comment} & \textbf{Label} \\
\hline
\rowcolor{red!15}
\textit{This is why we have 500+ million guns in america. 
We need to start using them.
If a man walked in my daughters bathroom, that would be his last day on earth.
People arent gonna keep tolerating this. 
Theres about to be rainbow street justice youre gonna start seeing.  When comes to my, or anyones kids, and i frankly dont care what your laws. Never did in the first place, but when it comes to my kid……ill wake up and chose violence EVERY time...} & Negative \\
\hline
\rowcolor{red!15}
 \textit{Hey if using the term f\censor{a}g is hate speech, then I stand guilty as charged. I hate what they stand for and I hate the corruption they're bringing into society, and I especially hate that they are preying on the minds of children  in an attempt to make them believe that their sick, perverted lifestyle is normal...}
 & Negative \\
\hline
\rowcolor{blue!15}
\textit{LISTEN PEOPLE: EVEN IF YOU DONT SUPPORT IT, YOU CAN RESPECT US AT LEAST. ALL WE WANT IS BASIC HUMAN DECENCY. LITERALLY JUST THAT. me and my girlfriend got spat at on a bus once, and then called homophobic slurs. we just want respect, being gay doesn't change the fact that I'm a human...}
& Positive \\
\hline
\rowcolor{blue!15}
\textit{Reading these comments are making me cry. I'm 13 and my Dad is transgender. 99\% of Doctors accept it. Why are you one to argue with Science? Please, if you are a bigot, don't reply to this comment. It will make me cry. I love my Dad, and you ignorant people who don't accept science won't change that....}& Positive \\
\hline
\end{tabular}
\caption{Illustrative examples of positive and negative YouTube comments identified by our classifier in the wild.}
\label{tab:youtube-comments}
\end{table*}


While hate speech detection and mitigation strategies have been extensively researched (see, e.g., \citealp{fortuna}), relatively little attention has gone into the opposite: \textit{hope speech}. Defined as comments and posts that inspire optimism and diffuse hostility in online spaces~\cite{palakodety2019hope,chakravarthi-2020-hopeedi}, hope speech is crucial for groups who face a disproportionate amount of hate speech.  In the current US cultural climate, where issues of race, gender, and sexuality are central to public debate~\cite{hartman2019war}, there is a need for tools that can detect both hope speech and negative content as listed in Table~\ref{tab:youtube-comments} and increase understanding of relevant online conversations. 
According to the FBI, one in five hate crimes in the US targets the LGBTQ+ community \cite{fbi}, and globally, homosexual activity remains punishable by death in six UN member states while only 37 countries recognize same-sex marriage \cite{ilga2024laws}. Understanding and improving online discourse around LGBTQ+ issues is thus a vital step toward protecting a vulnerable community.

\noindent\textbf{Contributions.} Our contributions are as follows.

\noindent\textcolor{blue}{(1)} In consultation with a health expert specializing in LGBTQ+ health for more than two decades, we curate a novel dataset of 3,750 instances with labels: \textit{neutral}, \textit{irrelevant}, \textit{positive} (hope speech), and \textit{negative}\footnote{The dataset is available at~\url{https://github.com/Social-Insights-Lab/LGBTQplus_HopeSpeech/}.}. Each instance is labeled by three raters (one Republican, one Democrat, and one independent) ensuring diverse and balanced political perspectives. 98 raters self-identified as being part of the LGBTQ+ community.\\
\textcolor{blue}{(2)} We analyze the association between rater political beliefs and how they rate content relevant to the LGBTQ+ community.\\
\textcolor{blue}{(3)} We analyze the alignment of LLMs with political beliefs in connection with LGBTQ+ content and how rater biases affect model fine-tuning.\\  
\textcolor{blue}{(4)} We provide novel insights into audience engagement patterns 
surrounding LGBTQ+ content, enhancing our
understanding of LGBTQ+ discussions within mainstream US political discourse. To our knowledge, no such study exists at our scale. 

\section{Related Work}

Our work is closely related with the \textit{hope speech} literature~\cite{palakodety2019hope,palakodety2020voice,chakravarthi-2020-hopeedi,yoo-etal-2021-empathy} and the broader literature of counter speech~\cite{benesch2014countering,mathew2019thou,DBLP:conf/naacl/HenglePSBA024,DBLP:conf/ijcai/SahaSKM022,DBLP:conf/acl/GuptaDGB0A23,DBLP:journals/corr/abs-2410-01400}. Our work contributes to this literature through our unique focus on LGBTQ+ discourse in mainstream US politics and our investigation on rater subjectivity and political leanings.

Political polarization in the US is widely studied across multiple disciplines in rich and diverse settings that include congressional voting patterns on policy issues~\cite{poole1984polarization}, mate selection~\cite{huber2017political}, allocation of scholarship funds~\cite{iyengar2015fear}, and annotating offensive content~\cite{sap-etal-2022-annotators,Vicarious}. Prior research showed systematic differences in offensive speech annotation based on annotators' beliefs~\cite{sap-etal-2022-annotators,Vicarious} and experiences~\cite{patton2019annotating}. While associations between political leanings and annotation of negative content has been studied in US politics~\cite{sap-etal-2022-annotators,Vicarious} and hot-button issues like reproductive rights and gun control~\cite{Vicarious}, our work extends this in two ways: first by examining perceptions of positive rather than negative content, and second by focusing specifically on LGBTQ+ discourse. Our study thus contributes to the broader literature of annotator subjectivity~\cite{pavlick2019inherent,sap2019risk, al2020identifying, larimore2021reconsidering, sap-etal-2022-annotators, goyal2022your, pei-jurgens-2023-annotator, Vicarious, homan2024intersectionality, prabhakaran2024grasp}. 

Our annotation study design is grounded in the prior literature~\cite{sap-etal-2022-annotators,Vicarious} and draws from~\citealp{Vicarious} and~\citealp{crowl2025measuring} in seeking diverse and balanced political perspectives including from the independents. Our work also touches upon political bias in LLMs as we observe that zero-shot classification of LGBTQ+ content of several models are more aligned with liberal raters~\cite{feng-etal-2023-pretraining,bang-etal-2024-measuring}. To address class imbalance, we construct an ensemble active learning pipeline much akin to~\citealp{palakodety2020voice} and~\citealp{IJCAIIran} leveraging well-known active learning strategies (e.g., certainty sampling~\cite{sindhwani2009uncertainty} and margin sampling~\cite{scheffer2001active}).

\begin{figure*}[t]
    \centering
    \includegraphics[width=0.8\textwidth]{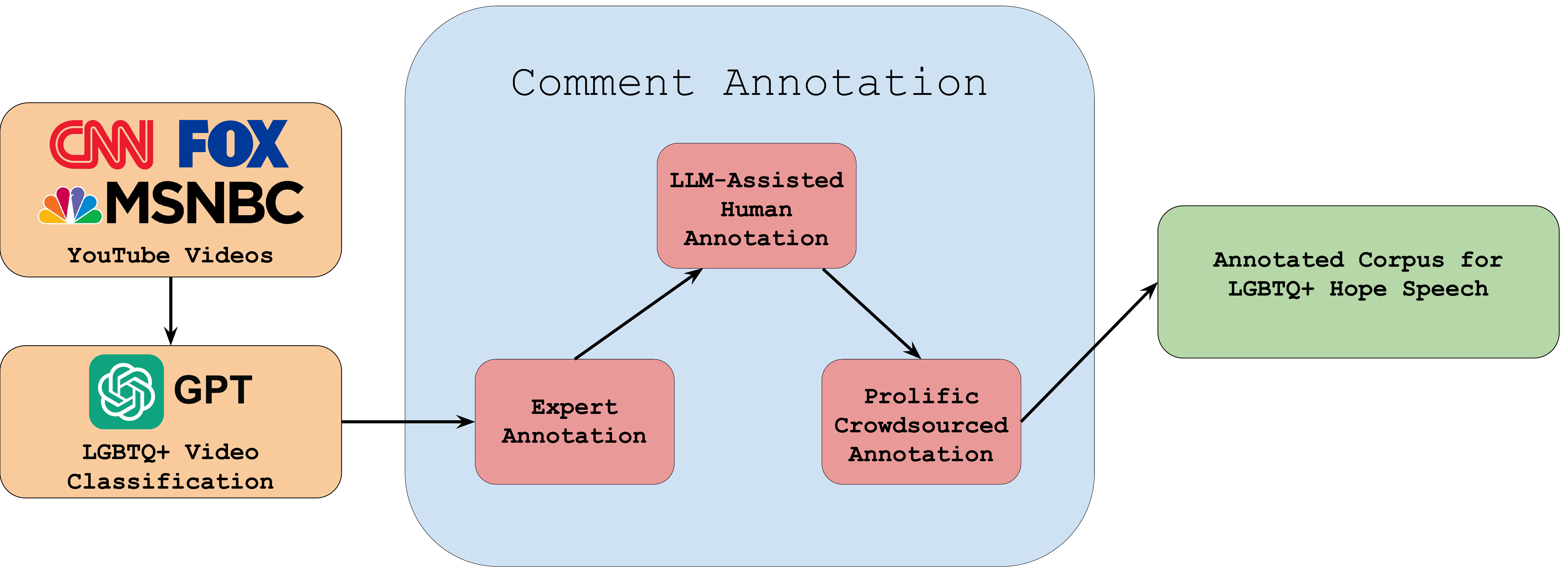}
    \caption{Corpus annotation pipeline}
    \label{fig:dataset}
\end{figure*}

\section{Dataset and Annotation Study Design}

Figure \ref{fig:dataset} presents a schematic diagram of our data collection process. which will be described in more detail in the following section.

\subsection{YouTube News Videos}

We consider a dataset of 333,032 YouTube videos uploaded between 04/10/2008 to 08/31/2024 by the official YouTube channels of three major US cable news networks: Fox News, CNN, and MSNBC. First introduced in~\citealp{KhudaBukhshPolarization}, this dataset represents a reliable snapshot of US political discourse encompassing diverse political perspectives over a substantial period of time. This dataset has found prior use in studying election-related discourse~\cite{Capitol,DBLP:conf/asunam/MittalCK24}, health-related discourse ~\cite{yoo2023auditingrobustifyingcovid19misinformation}, and rater subjectivity~\cite{Vicarious,DBLP:conf/emnlp/PanditaWDLRKZH24,ARTICLEAAAI2025}. We extend this dataset using the same data collection method to include videos up to August 2024.
\color{black}



\subsection{Identifying LGBTQ+ Relevant Videos}




We prune our initial set of videos using a two-step process. First, for each video, we pass the video title and description to \texttt{GPT-4o mini} prompting it to classify the video as specifcially relevant to the LGBTQ+ community or not  (Appendix contains prompt details in \S~\ref{sec:Prompt}). This step prunes our initial set of 333,032 videos down to 5,435 videos. Next, for each of these 5,435 videos, we use the same prompt and pass the video title and description to \texttt{GPT-4o}. While the \texttt{GPT-4o} model performed better and ended up with fewer false positives, we use this two-step process because it was considerably more expensive to run \texttt{GPT-4o} on all 333,032 videos. Overall, we obtain 3,161 relevant videos, denoted $\mathcal{V}_\textit{LGBTQ+}$. A random sample of 300 videos from this pipeline (100 from each channel) yielded 2 false positives for \emph{related} and no false negatives for \emph{unrelated}, indicating that our pruning steps are robust\footnote{Appendix contains examples (Table~\ref{tab:GPTExplanation}) of LLM explanations for a true positive and a false positive.}. 



\begin{table*}[htb]
\centering
\small
\setlength{\tabcolsep}{4pt}
\begin{tabular}{lcp{0.35\textwidth}rrrr}
\hline
\textbf{Dataset} & \textbf{Size} & \textbf{Sampling Strategy} & \textbf{Positive} & \textbf{Negative} & \textbf{Neutral} & \textbf{Irrelevant} \\
\hline
$\mathcal{D}_\textit{seed}$ & 1,950 & Initial crowd-sourced dataset& 649 (33.3\%) & 150 (7.7\%) & 87 (4.5\%) & 660 (33.8\%) \\
\hline
$\mathcal{D}_\textit{certainty}$ & 2,850 & Certainty sampling & 680 (23.9\%) & 682 (23.9\%) & 227 (8.0\%) & 764 (26.8\%) \\
\hline
$\mathcal{D}_\textit{margin}$ & 3,750 & Margin sampling~\cite{scheffer2001active}, a well-known variant of uncertainty sampling & 824 (22.0\%) & 947 (25.3\%) & 314 (8.4\%) & 1,020 (27.2\%) \\
\hline
$\mathcal{D}_\textit{eval}$ & 250 & Random set of comments taken from $\mathcal{D}_\textit{seed}$ held out from training for evaluation & 824 (22.0\%) & 947 (25.3\%) & 314 (8.4\%) & 1,020 (27.2\%) \\
\hline
\end{tabular}
\caption{{\small{Evolution of $\mathcal{D}_\textit{hope}$ through active learning stages, showing dataset size, sampling methodology, and label distribution.}}}
\label{tab:dataset-composition}
\end{table*}

\subsection{Constructing $\mathcal{D}_\textit{hope}$}

For each video in $\mathcal{V}_\textit{LGBTQ+}$, we collect user comments using publicly available YouTube API yielding 1,419,047 comments. From these comments, we construct $\mathcal{D}_\textit{hope}$, a dataset that categorizes content based on its stance toward the LGBTQ+ community. In consultation with the health expert specializing on LGBTQ+ health research for more than two decades, we identify four broad categories: \textit{positive} (\textit{hope speech}); \textit{negative}; \textit{irrelevant}; and \textit{neutral}. We define \textit{hope speech} as content that expresses support, advocacy, or acceptance for the LGBTQ+ community. \textit{Negative} content is defined as content that expresses opposition, discrimination, or hostility towards the LGBTQ+ community. \textit{Neutral} content is defined as content that is relevant to LGBTQ+ issues but does not belong to any of the previous two categories (\textit{hope speech} and \textit{negative}). \textit{Irrelevant} content is defined as content that is not related to LGBTQ+ issues. Detailed annotation guidelines are present in the Appendix (\S ~\ref{sec:guidelines}). 

In brief, this process consists of three steps:
\begin{compactenum}
\item We first construct a seed set of 1,950 instances using a collaborative human-LLM annotation framework. This seed set is annotated by crowd-sourced annotators for the aforementioned four categories  (details are described in \S~\ref{sec:human-LLM}). This step yields $\mathcal{D}_\textit{seed}$ consisting of 649 positives, 150 negatives, 87 neutrals, and 660 irrelevants and a validation set $\mathcal{D}_\textit{eval}$ of 250 instances set aside for all performance evaluation. 
\item Next, we conduct minority class certain sampling~\cite{sindhwani2009uncertainty} of additional 900 instances to address class imbalance (details are described in \S~\ref{sec:ActiveLearning}). This step yields $\mathcal{D}_\textit{certainty}$ consisting of 680 positives, 682 negatives, 227 neutrals, and 764 irrelevants. 
\item Finally, we conduct margin sampling~\cite{scheffer2001active} (described in \S~\ref{sec:ActiveLearning}) that yields $\mathcal{D}_\textit{margin}$ consisting of 824 positives, 947 negatives, 314 neutrals, and 1,020 irrelevants.   
\end{compactenum}


\subsubsection{Annotating $\mathcal{D}_\textit{hope}$}\label{sec:CrowdsourcedAnnotation}

We use Prolific for our annotation study, and hosted the survey on Qualtrics. Our study was reviewed by our Institutional Review Board and was deemed as exempt. The annotation study follows a similar annotation design as \citealp{Vicarious} and \citealp{crowl2025measuring} in considering raters evenly distributed across three political leanings (Republican, Democrat, and Independent). We split our corpus into a set of 65 batches, each containing an equal amount of comments from each channel. Each batch is labeled by three annotators (one Republican, one Democrat, and one Independent). Since we do not want to expose raters to objectionable content for a sustained period, following standard practices, we limit our batch size to 30 instances. We also left an unstructured text feedback input for each of the workers. While many raters thanked us for providing an interesting and thought-provoking task, some responded to the survey with homophobia (Appendix contains examples in \S~\ref{homophobia}) which might lead to interesting follow-on research. 

 On our final annotated dataset, we achieve moderate inter-rater agreement. The Fleiss' $\kappa$ for all four labels is 0.431 and for just two labels (\textit{HopeSpeech} (positive) and \textit{not-HopeSpeech} (neutral, negative, and irrelevant)) is 0.485. For a potentially subjective task like ours this agreement is in line with extensive prior literature. For instance, \citealp{guest-etal-2021} reported Fleiss’ $\kappa$ of 0.48 while \citealp{sanguinetti-etal-2018} reported category-wise $\kappa$ = 0.37. We further note that our observed inter-rater agreement is higher than \citealp{gomez-2020} ($\kappa$ = 0.15), \citealp{fortuna-2019} ($\kappa$ = 0.17) and \citealp{Vicarious} ($\kappa$ = 0.39). \citealp{Vicarious} studied political subjectivity in offensive content annotation and is closest to our study. 

\begin{table*}[htb]
\centering
\footnotesize
\setlength{\tabcolsep}{6pt}
\begin{tabular*}{\textwidth}{@{\extracolsep{\fill}}l|cccc|cccc@{}}
\hline
& \multicolumn{4}{c|}{\texttt{Llama}~\cite{dubey2024llama}} & \multicolumn{4}{c}{\texttt{Mistral}~\cite{jiang2023mistral}} \\
\textbf{Metric} & $\mathcal{M}(.)$ & $\mathcal{M}(\mathcal{D}_\textit{seed})$ & $\mathcal{M}(\mathcal{D}_\textit{certainty})$ & $\mathcal{M}(\mathcal{D}_\textit{margin})$ & $\mathcal{M}(.)$ & $\mathcal{M}(\mathcal{D}_\textit{seed})$ & $\mathcal{M}(\mathcal{D}_\textit{certainty})$ & $\mathcal{M}(\mathcal{D}_\textit{margin})$  \\
\hline
Accuracy & 0.5520 & 0.8480 & 0.8400 & 0.8480 & 0.6640 & 0.8120 & \textbf{0.8560} & 0.8480 \\
\hline
\multicolumn{9}{l}{\textbf{Macro-averaged metrics:}} \\
Precision & 0.5468 & \textbf{0.8025} & 0.7204 & 0.7993 & 0.5654 & 0.6297 & 0.7947 & 0.7960 \\
Recall & 0.5547 & 0.6366 & 0.6866 & \textbf{0.6947} & 0.5488 & 0.6001 & 0.6600 & 0.6843 \\
F1 Score & 0.4768 & 0.6680 & 0.6995 & \textbf{0.7162} & 0.5375 & 0.6017 & 0.6880 & 0.7087 \\
\hline
\multicolumn{9}{l}{\textbf{Class-specific F1 scores:}} \\
Positive & 0.6845 & 0.8703 & 0.8727 & 0.8661 & 0.8312 & 0.8448 & \textbf{0.8851} & 0.8722 \\
Negative & 0.3009 & 0.5806 & 0.5789 & \textbf{0.6842} & 0.4545 & 0.5714 & 0.5882 & 0.6486 \\
Neutral & 0.2553 & 0.2759 & \textbf{0.4103} & 0.3871 & 0.1765 & 0.0690 & 0.3333 & 0.3871 \\
Irrelevant & 0.6667 & \textbf{0.9453} & 0.9360 & 0.9275 & 0.6879 & 0.9216 & \textbf{0.9453} & 0.9268 \\
\hline
\end{tabular*}
\caption{Comparison of \texttt{Llama} and \texttt{Mistral} model results on aggregate labels over five independent training runs on {$\mathcal{D}_\textit{eval}$. $\mathcal{M}(.)$} \color{black} denotes the zero-shot performance. $\mathcal{M}(\mathcal{D})$ denotes model performance when fine-tuned on $\mathcal{D}$. Table \ref{tab:dataset-composition} summarizes dataset statistics.}
\label{tab:model-comparison}
\end{table*}

\section{Classification Results}

We consider \texttt{Mistral} (Mistral-7B-Instruct-v0.3)~\cite{jiang2023mistral} and \texttt{Llama 3} (Llama-3-8B-Instruct)~\cite{dubey2024llama}, two well-known open LLMs for fine-tuning.       
We determine the aggregate label by taking a majority vote, a standard approach to resolving annotator disagreement (see, e.g.,~\citealp{davidson2017automated,wiegand2019detection}). Any comments that received three separate labels are not used in the aggregate data, accounting for 11.3\% of instances. In addition to fine-tuning models on aggregate labels, we also fine-tune models specific to each political affiliation. For those models, we use these 11.3\% additional instances\footnote{Table~\ref{tab:political-agreement-finetuned} describes the noise audit results of these models.}.


Table~\ref{tab:model-comparison} summarizes our models' performance on $\mathcal{D}_\textit{eval}$. Collectively, we refer to our annotated corpus as $\mathcal{D}_\textit{hope}$, which evolved through three active learning phases as seen in Table \ref{tab:dataset-composition}: initial seed collection ($\mathcal{D}_\textit{seed}$) comprising 1,950 instances, certainty sampling ($\mathcal{D}_\textit{certainty}$) which expanded the dataset to 2,850 instances, and margin sampling ($\mathcal{D}_\textit{margin}$) which yielded our final dataset of 3,750 instances. $\mathcal{D}_\textit{eval}$ consists of 250 instances randomly selected from $\mathcal{D}_\textit{seed}$ and held out from all training processes, serving as a consistent evaluation set for all models. \color{black} We consider both zero-shot and fine-tuned models (Appendix contains prompt details in \S~\ref{sec:Prompt}). For fine-tuning, we use a sequence classification head. We split the training data, after removing the validation set, into an 80/20 test/train split and train for 5 epochs, considering the best model as the one with the highest Macro-F1. We find (1) the fine-tuned models considerably outperform zero-shot models; (2) the models' performance improve  gradually across different stages of the active learning pipeline; and (3) the best-performing model is \texttt{Llama 3} trained on $\mathcal{D}_\textit{margin}$.

For the sake of completeness, we compare performance against several off‐the‐shelf classifiers on adjacent tasks such as hate speech detection, sentiment classification, and stance detection (results in Table \ref{tab:baseline_comparison}). As sentiment classifier baselines, we consider a \texttt{DistilBERT} model~\cite{sanh2019distilbert} fine-tuned on SST-2~\cite{socher-etal-2013-recursive}, which provides only positive/negative predictions. For fair comparison, we map positive sentiment to \textit{hope speech} and negative sentiment to \textit{negative} content, while removing any comments with \textit{neutral} or \textit{irrelevant} content. It achieves a Macro F1 of 0.601 and Positive F1 of 0.520, significantly lower than our best \texttt{Llama}-based model (Macro F1: 0.864, Positive F1: 0.685), even with the removal of neutral/irrelevant label noise. \color{black} This gap stems from sentiment classifiers failing to capture intent in LGBTQ+ discourse. For instance, a sentiment classifier will predict \textit{I will be extremely happy if Congress passes a resolution and takes away voting rights from the LGBTQ+ people} as positive. 

We further compare our approach with toxicity and hate speech detectors. We use \texttt{Detoxify}~\cite{Detoxify} for toxic comment detection and cardiffnlp’s \texttt{hate-roBERTa}~\cite{barbieri-etal-2020-tweeteval} for hate speech detection. For fair comparison, we map toxic and hateful content to negative speech and hope, neutral, or irrelevant content to non–toxic and non-hateful speech. Therefore, we report their Negative F1 scores (i.e., their ability to correctly identify harmful content) rather than Positive F1. These models achieved low Negative F1 scores (0.129 and 0.258 respectively), as they focus on overt toxicity, while in LGBTQ+ discourse disapproval can often manifest without explicit hate speech.

Finally, our task bears many similarities to that of determining stance on the LGBTQ+ community. However, off-the-shelf stance detectors are typically developed for specific topics (e.g., abortion or climate change), and there are few systems available that allow for input of a custom topic beyond using LLMs to specify in the prompt. In practice, these LLM-based approaches for stance detection end up resembling the zero-shot evaluations we have already reported.  

\noindent\textbf{Error analysis.} Consistent with the existing literature~\cite{joshi2017automatic,DBLP:conf/ijcai/FarabiRKKZ24}, we observe that our models often struggled in detecting sarcasm. Appendix (\S~\ref{error-analysis}) discusses challenging examples.




\section{Substantive Findings}


\subsection{User Engagement Findings}

$\mathcal{V}_\textit{LGBTQ+}$ represents a set of 3,161 news videos relevant to LGBTQ+ issues. We first construct a control set, $\mathcal{V}_\textit{control}$. Each video $\textit{v}$ in $\mathcal{V}_\textit{LGBTQ+}$ is matched with a video that is (1) not relevant to LGBTQ+ issues; (2) is uploaded by the same news outlet; (3) has roughly same video duration as $\textit{v}$; and (4) has the publishing time closest to $\textit{v}$.   

Let for a given video $v$, $v_\textit{likes}$, $v_\textit{dislikes}$, $v_\textit{views}$, and $v_\textit{comments}$ denote the total number of likes, dislikes, views and comments on $v$, respectively. 
Following~\citealp{Lopez-youtubers,Entube,braun2017enhanced,KhudaBukhshPolarization}, we calculate engagement rate and dislike ratio for a given video as follows.\\
\textit{engagementRate}($v$) = $\frac{v_\textit{likes} + v_\textit{dislikes} + v_\textit{comments}}{v_\textit{views}}$\\
\textit{dislikeRate}($v$) = $\frac{v_\textit{dislikes}}{v_\textit{dislikes} + v_\textit{likes}}$

\normalsize

\begin{table}[htb]
\centering
\footnotesize
\begin{tabular}{@{}llcccc@{}}
\hline
\textbf{Outlet} & \textbf{Type} & \multicolumn{2}{c}{\textbf{Dislike Rate}} & \multicolumn{2}{c}{\textbf{Engagement Rate}} \\
 & & \textbf{Mean} & \textbf{Med.} & \textbf{Mean} & \textbf{Med.} \\
\hline
\multirow{2}{*}{MSNBC} & $\mathcal{V}_\textit{control}$ & 22.7\% & 13.5\% & 2.69\% & 2.47\% \\
 & $\mathcal{V}_\textit{LGBTQ+}$ & 24.1\% & 18.8\% & 2.60\% & 2.30\% \\
\hline
\multirow{2}{*}{FOX} & $\mathcal{V}_\textit{control}$ & 18.7\% & 10.6\% & 3.40\% & 3.29\% \\
 & $\mathcal{V}_\textit{LGBTQ+}$ & 29.7\% & 20.8\% & 2.63\% & 2.52\% \\
\hline
\multirow{2}{*}{CNN} & $\mathcal{V}_\textit{control}$ & 27.3\% & 21.2\% & 1.69\% & 1.31\% \\
 & $\mathcal{V}_\textit{LGBTQ+}$ & 35.5\% & 32.4\% & 1.66\% & 1.30\% \\
\hline
\end{tabular}
\caption{\small{Dislike and Engagement Rates by outlet and video type.}}
\label{tab:dislike-engagement-rates}
\end{table}

Table \ref{tab:dislike-engagement-rates} compares the engagement rate and dislike rates of videos in $\mathcal{V}_\textit{LGBTQ+}$ and $\mathcal{V}_\textit{control}$. YouTube stopped allowing  users to access dislikes starting in 2021, so our dislike rate data only includes videos up until that point\footnote{sourced from~\citealp{dutta2024anonymous}}. We observe that LGBTQ+ videos consistently received higher dislike rates across all three channels compared to non-LGBTQ+ content. This difference is most pronounced for FOX News and CNN, with smaller but still noticeable differences for MSNBC. Overall engagement rates for LGBTQ+ videos are generally lower than for non-LGBTQ+ content, though the magnitude of this difference varied by news outlet. FOX News showed the largest decrease in engagement for LGBTQ+ content, while CNN and MSNBC exhibited differences small enough to likely be insignificant. The higher dislike ratios and lower engagement rate for LGBTQ+ content across all news outlets, regardless of perceived political leaning, point to a persistent societal resistance to LGBTQ+ topics. Although, this resistance is shown to be more pronounced for the most conservative channel, FOX News. Appendix \ref{sec:controversial-topics} and \ref{sec:video-content} contain additional contrastive analyses with other hot-button issues and a deeper look into heavily disliked videos.   


\subsection{Annotation Study Findings}

Table~\ref{tab:political-agreement} suggests partisan differences in the annotation of \textit{hope speech} for the LGBTQ+ community. Even given the same comment text and annotation guidelines, Republicans and Democrats or Independents may have different conclusions as to what is hope speech or negative towards the LGBTQ+ community. The agreement between Democrats and Independents is higher than the agreement between either group and Republicans. 

\par Our survey also includes a question\footnote{Raters could opt out of answering this question.} asking the raters if they considered themselves a member of the LGBTQ+ community. Out of 375 raters, 98 (31 Democrats; 29 Republicans; and 38 Independents) identified themselves as being LGBTQ+. The Cohen's $\kappa$ agreement between LGBTQ+ and non-LGBTQ+ annotators is 0.449. This agreement fits between the low values found when comparing Republicans to others, and the high value found comparing Independents and Democrats and is close to the overall Fleiss' $\kappa$ of 0.431. We find this result particularly interesting as it indicates that even for a topic highly relevant to gender and sexual orientation, rater disagreement is perhaps impacted more by political leanings than sexual orientation. Even if we collapse the labels into \textit{HopeSpeech} and \textit{not-HopeSpeech}, the qualitative findings on political disagreement remain unchanged (see, Appendix).

To better understand intra-group annotation agreement, we collect additional annotations for a subset of 30 batches where each instance was labeled by two annotators of the same political affiliation. Independent annotators show the highest internal consistency ($\kappa$ = 0.500), suggesting more unified perspectives within this group than the Democrats ($\kappa$ = 0.446) and the Republicans ($\kappa$ = 0.368). The Republicans show the lowest intra-group agreement (even lower than their inter-group agreement with Democrats or Independents) suggesting more varied perspectives within the group.

To analyze the intersection of LGBTQ+ identity and political beliefs, we next conduct two-proportion z-tests comparing label distributions between LGBTQ+ and non-LGBTQ+ annotators within each political affiliation. The analysis has two key major takeaways (details in Table~\ref{tab:lgbtq-analysis}). First, LGBTQ+ annotators across all political affiliations were consistently less likely to label content as irrelevant compared to their non-LGBTQ+ counterparts (Democrats: -4.1\%, \textit{p} < 0.05; Republicans: -6.0\%, \textit{p} < 0.001; Independents: -3.5\%, \textit{p} < 0.05), suggesting they more readily recognize content relevant to LGBTQ+ issues. Table \ref{tab:human-examples} shows a comment that a LGBTQ+ annotator correctly identified as LGBTQ+-related, while the two other annotators seemed to have not enough context to make the connection. Second, the strongest differences appear among Republican annotators, where LGBTQ+ Republicans were significantly more likely to label content as negative  and less likely to label it as positive compared to non-LGBTQ+ Republicans. These findings suggest that lived experience as an LGBTQ+ person may affect content interpretation, with the strongest differences emerging for ideologically charged content among conservative annotators.

\color{black}

\begin{table}[t]
\centering
\small
\setlength{\extrarowheight}{2pt}
\begin{tabular}{cc|c|c|c|}
  & \multicolumn{1}{c}{} & \multicolumn{1}{c}{\textit{Dem}}  & \multicolumn{1}{c}{\textit{Rep}}  & \multicolumn{1}{c}{\textit{Ind}} \\\cline{3-5}
            & \textit{Dem} &\cellcolor{blue!25} - & 0.408  & 0.477
 \\ \cline{3-5}
 & \textit{Rep} & \textcolor{black}{0.408} &\cellcolor{blue!25} - & \textcolor{black}{0.396}
 \\\cline{3-5}
            & \textit{Ind} & \textcolor{black}{0.477} & \textcolor{black}{0.396 }  &\cellcolor{blue!25} - \\\cline{3-5}
\end{tabular}
\caption{Human rater agreement across political affiliations. A cell $\langle i,j\rangle$ presents the Cohen's $\kappa$ agreement between raters with political affiliation $i$ and $j$. \textit{Dem}, \textit{Rep}, and \textit{Ind} denote Democrat, Republican, and Independent, respectively.}
\label{tab:political-agreement}
\end{table}


In addition to our main model fine-tuned on the aggregate labels, we also fine-tune models specifically on the labels of each political affiliation (denoted by $\mathcal{M}_\textit{rep}$, $\mathcal{M}_\textit{dem}$, and $\mathcal{M}_\textit{ind}$).  $\mathcal{M}_\textit{rep}$ was trained only on the 3,750 comments labeled by Republicans, $\mathcal{M}_\textit{dem}$ only on the 3,750 labeled by Democrats, and $\mathcal{M}_\textit{ind}$ on the 3,750 labeled by Independents. Using each of these models, we classify a set of 50k unseen comments from each channel. 
Following~\cite{kahneman2021noise,Vicarious}, we conduct a \textit{noise audit} of these models as shown in Table \ref{tab:political-agreement-finetuned}. A \textit{noise audit} involves examining agreement between different model variants on identical inputs, revealing systematic differences in model predictions. 
To verify whether political biases are introduced during fine-tuning, we first examine agreement between pre-trained and fine-tuned models. We observe notably low agreement scores: Cohen's $\kappa$  of 0.29 for both $\mathcal{M}_\textit{ind}$ and $\mathcal{M}_\textit{dem}$ with pre-trained \texttt{Llama}, and 0.25 for $\mathcal{M}_\textit{rep}$ with pre-trained \texttt{Llama}. These low agreement scores suggest substantial shifts in model behavior during fine-tuning.
While we observe the in-the-wild agreement across models is higher than the agreement across human raters with different political beliefs, the impact of the political affiliations still shone through. Once again, $\mathcal{M}_\textit{dem}$ and $\mathcal{M}_\textit{ind}$ share higher agreement with each other, while they both have lower agreement with $\mathcal{M}_\textit{rep}$ to a similar degree. This demonstrates that the human biases due to political affiliation in labeled data can perpetuate to fine-tuned models. Tables~\ref{tab:political-commentsHumans} and~\ref{tab:political-comments} present illustrative examples showing how trans-exclusionary stance of human Republican raters may get perpetuated to $\mathcal{M}_\textit{rep}$ affecting marginalized voices. 


\begin{table}[htb]
\centering
\small
\begin{tabular}{|p{0.6\linewidth}|p{0.12\linewidth}|p{0.12\linewidth}|}
\hline
\textbf{Comment} & \textit{Rep} & \textit{Dem} \& \textit{Ind} \\
\hline
\textit{Good Job America - Keep it UP- GO WOKE GO BROKE - Transheuser-Busch deserves IT} &
Positive &
Negative \\
\hline
\textit{Thank you so much for sharing. Your story highlights how messed up the treatment is for Trans people.} &
Negative &
Positive \\
\hline
\end{tabular}
\caption{Disagreements in labels across human raters on Prolific. \textit{Rep}, \textit{Dem}, and \textit{Ind} denote Republican, Democrat, and Independent, respectively.}
\label{tab:political-commentsHumans}
\end{table}

\begin{table}[h]
\centering
\small
\begin{tabular}{|p{0.6\linewidth}|p{0.12\linewidth}|p{0.12\linewidth}|}
\hline
\textbf{Comment} & $\mathcal{M}_\textit{rep}$ & $\mathcal{M}_\textit{dem} $, $\mathcal{M}_\textit{ind}$ \\
\hline
\textit{Since when was having a phobia such a bad thing? You wouldnt say this about some scared of heights of or spiders… there freaks amd theres nothing to be ashamed of when having a transphobia} &
Positive &
Negative \\
\hline
\textit{damn! first time I've seen Tucker actually lose. I love you Tucker, but the military transgender ban is BS. it's not a huge expense. Like Sean Patrick said in the end before Tucker rage quits "They're earning those benefits by putting their life on the line for America and have done more for this country than you (Tucker) and I ever will"...} &
Negative &
Positive \\
\hline
\end{tabular}
\caption{Illustrative examples where $\mathcal{M}_\textit{rep}$ does not agree with $\mathcal{M}_\textit{dem}$ and $\mathcal{M}_\textit{ind}$. }
\label{tab:political-comments}
\end{table}

\begin{table}[t]
\centering
\small
\setlength{\extrarowheight}{2pt}
\begin{tabular}{cc|c|c|c|}
  & \multicolumn{1}{c}{} & \multicolumn{1}{c}{$\mathcal{M}_\textit{dem}$}  & \multicolumn{1}{c}{$\mathcal{M}_\textit{rep}$}  & \multicolumn{1}{c}{$\mathcal{M}_\textit{ind}$} \\\cline{3-5}
            & $\mathcal{M}_\textit{dem}$ &\cellcolor{blue!25} - & 0.572  & 0.670
 \\ \cline{3-5}
 & $\mathcal{M}_\textit{rep}$ & \textcolor{black}{0.572} &\cellcolor{blue!25} - & \textcolor{black}{0.573}
 \\\cline{3-5}
            & $\mathcal{M}_\textit{ind}$ & \textcolor{black}{0.670} & \textcolor{black}{0.573}  &\cellcolor{blue!25} - \\\cline{3-5}
\end{tabular}
\caption{{In-the-wild Cohen’s $\kappa$ agreement of models trained on data annotated by raters with same political affiliation.}}
\label{tab:political-agreement-finetuned}
\end{table}

We also calculate the performance metrics of zero-shot classification against the annotations of each political affiliation (see, Table \ref{tab:political-affiliation-comparison}). The results show similar performance with Democrat and Independent labels, but each metrics was lower for the Republican labels. Similar trend was present across the three different models we tested; \texttt{GPT-4o-mini}, \texttt{Llama}, and \texttt{Mistral}. This suggests that the biases present in these models may align closer with the biases found in our human Independent and Democrat raters, than those of a Republican.

In addition, we conduct a parallel analysis training separate models on annotations from LGBTQ+ and non-LGBTQ+ annotators. For fair comparison, in this experiment, we use a reduced dataset where comments had labels from both LGBTQ+ and non-LGBTQ+ annotators. Consistent with our findings from human annotators, we observe that the non-LGBTQ+ model more frequently classified content as "irrelevant" compared to the LGBTQ+ model. Table~\ref{tab:human-examples} shows an example where the LGBTQ+ model identified as relevant content that non-LGBTQ+ annotators marked as irrelevant.
\color{black}
\begin{table*}[t]
\centering
\setlength{\tabcolsep}{3pt}
\footnotesize
\begin{tabular*}{\textwidth}{@{\extracolsep{\fill}}l|ccc|ccc|ccc@{}}
\hline
\multirow{2}{*}{\textbf{Metric}} & \multicolumn{3}{c|}{\textbf{GPT-4o-mini}} & \multicolumn{3}{c|}{\textbf{Meta-Llama-3-8B}} & \multicolumn{3}{c}{\textbf{Mistral-7B-v0.3}} \\
\cline{2-10}
 & \textbf{Dem.} & \textbf{Rep.} & \textbf{Ind.} & \textbf{Dem.} & \textbf{Rep.} & \textbf{Ind.} & \textbf{Dem.} & \textbf{Rep.} & \textbf{Ind.} \\
\hline
Accuracy & \textbf{0.5963} & 0.5520 & 0.5864 & 0.5143 & 0.4809 & 0.5091 & 0.5354 & 0.4954 & 0.5260 \\
\hline
\multicolumn{10}{l}{\textbf{Macro-averaged metrics:}} \\
Precision & \textbf{0.6146} & 0.5664 & 0.6057 & 0.4534 & 0.4101 & 0.4484 & 0.4316 & 0.4019 & 0.4236 \\
Recall & \textbf{0.5839} & 0.5391 & 0.5717 & 0.3803 & 0.3497 & 0.3736 & 0.4178 & 0.3849 & 0.4061 \\
F1 Score & \textbf{0.5749} & 0.5320 & 0.5636 & 0.3716 & 0.3401 & 0.3653 & 0.4080 & 0.3790 & 0.3990 \\
\hline
\multicolumn{10}{l}{\textbf{Class-specific F1 scores:}} \\
Positive & \textbf{0.7236} & 0.6464 & 0.7113 & 0.5801 & 0.5102 & 0.5609 & 0.6586 & 0.5884 & 0.6469 \\
Negative & \textbf{0.6509} & 0.6171 & 0.6330 & 0.5846 & 0.5670 & 0.5781 & 0.6103 & 0.5571 & 0.5886 \\
Neutral & \textbf{0.3406} & 0.3233 & 0.3264 & 0.2096 & 0.1718 & 0.1943 & 0.2604 & 0.2537 & 0.2453 \\
Irrelevant & \textbf{0.5843} & 0.5412 & 0.5840 & 0.4840 & 0.4513 & 0.4930 & 0.5108 & 0.4957 & 0.5141 \\
\hline
\end{tabular*}
\caption{{Zero-shot model results across political affiliation specific labels evaluated on the full human annotated corpus of 3,750 comments.}}
\label{tab:political-affiliation-comparison}
\end{table*}

\noindent\textbf{Key takeaways:} \textcolor{blue}{(1)} Annotation of \textit{hope speech} is associated with political leanings that drive more disagreement than gender and sexual identity on a task relevant to gender and sexual identity; \textcolor{blue}{(2)} LGBTQ+ annotators across all political affiliations were consistently less likely to label content as irrelevant and showed stronger recognition of LGBTQ+-related content, with the most pronounced differences appearing among Republican annotators; \textcolor{blue}{(3)} political biases in human annotations carried over to fine-tuned language models, with models trained on Democrats and Independents showing higher agreement compared to the Republican trained model. 

\subsection{In-the-wild \textit{hope speech} findings}

In order to look at the hope and hate speech make-up of LGBTQ+ discussion in-the-wild, we consider our best-performing model and classify 50k randomly sampled, unseen comments from each news outlet. The results, as shown in Figure \ref{fig:label-breakdown-with-ci} with numerical results in Table \ref{tab:in-the-wild}, reveal distinct patterns. In particular, FOX News has the highest proportion of negative comments (24.15\%) and the lowest proportion of \textit{hope speech} (2.3\%) among the three outlets. This aligns with the outlet's reputation for conservative viewpoints. In contrast, CNN shows a more balanced distribution, with the highest percentage of \textit{hope speech} (7.66\%) and a negative comment rate (17.2\%) closer to the overall average. MSNBC, however, has a much lower proportion of \textit{hope speech} than CNN (4.55\%) and the lowest proportion of negative comments (9.36\%).
 \par 
Figure~\ref{fig:label-breakdown-with-ci} reports the ratio of \textit{hope speech} and \textit{hope speech} and \textit{negative}s found in the wild. Dubbed \textit{positivity ratio}, this ratio focuses on comments expressing polarized stance towards the LGBTQ+ community, excluding \textit{irrelevant} and \textit{neutral} comments. We find that MSNBC has the highest \textit{positivity ratio} at 32.7\%, followed by CNN at 30.81\%, both above the overall average of 22.25\%. In stark contrast, FOX News shows a considerably lower \textit{positivity ratio} of 8.7\%. These findings further emphasize the divergence in audience sentiment across channels, with MSNBC and CNN fostering more balanced or slightly supportive discussions around LGBTQ+ content, while FOX News comments lean heavily towards negative sentiment. To better understand the nature of these discussions beyond labels, Appendix \ref{sec:lexical-analysis} contains a lexical analysis of the language used in different comment categories.


\begin{figure}[htbp]
\centering
\begin{tikzpicture}
\begin{axis}[
    width=\columnwidth,
    height=0.6\columnwidth,
    ybar,
    bar width=7pt,
    enlarge x limits={0.15},
    legend style={at={(0.5,1.05)}, anchor=south, legend columns=-1},
    ylabel={Percentage},
    symbolic x coords={Irrelevant, Negative, Positive, Neutral},
    xtick=data,
    x tick label style={rotate=45,anchor=east},
    ymin=0,
    ymax=100,
    scaled y ticks = false,
    ylabel near ticks,
    title style={font=\large},
    label style={font=\small},
    tick label style={font=\small},
    y label style={at={(axis description cs:-0.1,.5)}, anchor=south},
    error bars/y dir=both,
    error bars/y explicit,
    error bars/error bar style={line width=0.75pt, color=black},
]
\legend{Overall, CNN, FOX, MSNBC}
\addplot[fill=green!40, error bars/.cd, y dir=both, y explicit] 
    coordinates {
        (Irrelevant, 76.53) +- (0.25, 0.25)
        (Negative, 16.90) +- (0.25, 0.25)
        (Positive, 4.84) +- (0.15, 0.15)
        (Neutral, 1.73) +- (0.1, 0.1)
    };
\addplot[fill=blue!40, error bars/.cd, y dir=both, y explicit] 
    coordinates {
        (Irrelevant, 72.96) +- (0.6, 0.6)
        (Negative, 17.20) +- (0.5, 0.5)
        (Positive, 7.66) +- (0.35, 0.35)
        (Neutral, 2.19) +- (0.2, 0.2)
    };
\addplot[fill=red!40, error bars/.cd, y dir=both, y explicit] 
    coordinates {
        (Irrelevant, 71.78) +- (0.45, 0.45)
        (Negative, 24.15) +- (0.45, 0.45)
        (Positive, 2.30) +- (0.2, 0.2)
        (Neutral, 1.77) +- (0.2, 0.2)
    };
\addplot[fill=orange!40, error bars/.cd, y dir=both, y explicit] 
    coordinates {
        (Irrelevant, 84.87) +- (0.45, 0.45)
        (Negative, 9.36) +- (0.4, 0.4)
        (Positive, 4.55) +- (0.2, 0.2)
        (Neutral, 1.22) +- (0.1, 0.1)
    };
\end{axis}
\end{tikzpicture}
\caption{Label breakdown by news outlets with 95\% confidence intervals. 50k comments from each news outlet found in-the-wild were classified by our best performing model. Numerical values are listed in Table~\ref{tab:in-the-wild}.}
\label{fig:label-breakdown-with-ci}
\end{figure}

\par
Perhaps the most important results come from looking at the trends as a whole. For every channel, there were more negative comments than \textit{hope speech}. This consistent imbalance, suggests a broader societal tendency towards critical or oppositional engagement with LGBTQ+ topics in online spaces. It implies that, regardless of the audience, viewers are more inclined to express disapproval or criticism than support or affirmation when commenting on LGBTQ+-related content.  \footnote{Our Limitations and Appendix contain an exploratory study into Reddit showing that there are certain social web pockets which may be more supportive than negative, warranting additional investigation into more varied social media.}
Another noteworthy trend is the uniformly low percentage of neutral comments across all channels, ranging from 1.22\% to 2.22\%. This scarcity of neutral perspectives, coupled with the positive-negative imbalance, points to a highly polarized discourse surrounding LGBTQ+ issues. It suggests that those who engage in these comment sections tend to hold and express strong opinions, whether positive or negative, rather than maintaining a neutral stance. 

To confirm the accuracy of our model when generalizing to in-the-wild data, we conducted a manual inspection of 100 comments labelled as Positive. From those 100, we found 16 false positives, mainly consisting of comments that we would label as Negative according to our guidelines, such as:
\begin{quote}
LGBTQHDJFNEIEKBEJDJDNDIDBDJ\-IDJDIGWJOQPQPLAIDHWNRBHXKS\-OFORBRIDKWNDIKFKDNRUWODUE
Get it together bigot!
\end{quote}

This example shows a comment that is Negative towards the LGBTQ+ community in a more subtle or sarcastic way. An error analysis in Appendix \ref{error-analysis} further discusses some limitations that we identified through challenging comments it labeled incorrectly.
\color{black}

\section{Conclusions}

This paper presents a comprehensive picture of LGBTQ+ discussions in mainstream US political discourse. Overseen by a health expert specializing in LGBTQ+ health for decades, we conduct a detailed annotation study that reveals political biases of raters are associated with how they rate supportive content for the LGBTQ+ community. Our study shows that such biases may perpetuate into models affecting marginalized voices. Our in-the-wild assessment of LGBTQ+ discussions reveal that negative comments about the community considerably outweigh positive discourse indicating a technological gap to ensure safer spaces for marginalized communities.   

\clearpage
\section{Ethics Statement}

Our study was reviewed by our Institutional Review Board and was deemed as exempt. Our study is overseen by a health expert with decades of research on LGBTQ+ health. We investigate publicly available data collected using public APIs. 

We do not collect any PII (Personally Identifiable Information) about the raters and compensate them above the minimum wage. Since content moderation can be potentially gruesome and affect the
mental health of the raters~\cite{Guardian1}, we
maintain a small batch size (30 YouTube comments).

While our goal
is to broaden our understanding of LGBTQ+ discussions in mainstream US political news and our content classifier can assist human moderators to identify supportive content for the LGBTQ+ community, any content
filtering system can be tweaked for malicious purposes. For instance, an inverse filter can be made
that filters out \textit{hopeSpeech} posts while filtering in \textit{not-hopeSpeech} 
ones.

Our substantive findings rely on fine-tuned LLM. Studies indicate that these models have a
wide range of biases that reflect the texts on which they were
originally trained, and which may percolate to downstream
tasks~\cite{bender2021dangers}. 

\section{Limitations}
While our study offers valuable insights into hope speech detection for LGBTQ+ topics in US political discourse, we recognize there are limitations that shape the scope and applicability of the findings.
\par
Our focus on YouTube comments for major news channels, while providing a rich dataset, is only a fraction of the online discource related to LGBTQ+. Platforms like Twitter, Reddit, and others have their own unique demographics, content, and user interactions. What we observed on YouTube may not be true for the rest of the Internet, potentially limiting our generalizability across social media. We conducted an exploratory investigation on Reddit, sampling 15,000 comments on LGBTQ+-relevant posts (2016-2024) from r/politics, r/republican, and r/democrat, for a total of 45,000 comments. Overall, the positivity ratio remains below 0.5 (0.40), meaning that negative comments outnumber hope speech on Reddit, as on YouTube.  However the subreddit-level breakdown is more nuanced, the r/politics and r/democrat subreddits show a slight tilt toward hope speech, whereas r/republican is strongly negative. This differs from YouTube, where even communities one might expect to have more positivity, such as the CNN comment section, still had negative comments outnumbering hope speech. These results, seen in Table \ref{tab:subreddit-distribution},  underline the fact that future work across additional communities and media is necessary. \color{black}
\par Additionally, our study is grounded in US politics, where societal attitudes towards the LGBTQ+ have been affected by historical and legal context. However, LGBTQ+ rights and acceptance may vary dramatically across the world. In some countries, open support for LGBTQ+ rights might be more common and less contentious, while in others, it could be far more risky or even illegal to express such support. This global variation in LGBTQ+ rights and societal attitudes means that the patterns of hope speech and the very definition of what constitutes supportive language could differ significantly across cultural and national boundaries. Recent research \cite{samir2024locatinginformationgapsnarrative} demonstrates these cross-cultural differences concretely, finding significant variations in how LGBTQ+ individuals are portrayed across different language Wikipedias, with some versions systematically emphasizing negative biographical information or omitting positive achievements based on local cultural attitudes.

\par While our study did not specifically aim at participatory AI~\cite{harrington2019deconstructing,birhane2022power} as our key goal was to study the interplay of rater politics and how they perceive LGBTQ+ discussions, a considerable fraction of our raters self-reported as being part of the community. Future studies can solely focus on participatory AI involving more raters from the LGBTQ+ community and extend this research to vicarious interactions~\cite{Vicarious}. 

\par We also acknowledge that many challenges faced by individual groups within the LGBTQ+ community are unique. For instance, the trans community faces several additional challenges such as participation in competitive sports or access to gender reassignment treatment. Future studies solely focusing on the trans community will add further value to prior literature focusing on specific exclusionary behavior (see, e.g.,~\citealp{lu-jurgens-2022-subtle}). 
\par In terms of political affiliations, while we attempted to capture the broad ideological difference present in the US, this may oversimplify their nuanced reality. Our categorization of political affiliations into the groups of "Democrat," "Republican," and "Independent" may not fully capture these nuances. In reality, political views exist on a spectrum rather than in discrete categories. 
\par While we have attempted to make the definition of hope speech clear, operationalizable, and indicate how it differs from positive sentiment, we acknowledge room for further nuance. Appendix~\ref{sec:guidelines} lists a nuanced definition of \textit{hope speech} which we considered in our initial annotation phase conducted by the author. However, our initial pilot revealed that this fine-grained definition resulted in poor quality annotation and low annotator agreement. Keeping rater instructions simple to reduce cognitive load on raters is a recommended practice in the crowd-sourcing literature~\cite{finnerty2013keep}. To reduce cognitive load, we adopted a simpler four-way scheme, \textit{Positive}, \textit{Negative}, \textit{Neutral}, \textit{Irrelevant}, and worded our annotation guidelines as advised by the subject-matter expert using the more familiar term \textit{Positive} in place of \textit{Hope Speech}. Although this worked well for the purposes of our study, a finer-grained taxonomy and clearer operational guidelines merit deeper investigation. \color{black}
\par Finally, we acknowledge a lack of intersectionality analysis in our study. We have focused on LGBTQ+ identity and political affiliation, overlooking the possibility of intersection with factors such as race, ethnicity, age, or socioeconomic status. The interplay of these identities can have significant influence on how an individual perceives the LGBTQ+ community and speech surrounding it. Future work could be improved by capturing these complex intersection and identifying possible impact on LGBTQ+ and hope speech perception.

\section{Acknowledgements}
We acknowledge Research Computing at the Rochester Institute of Technology for providing computational resources. We thank Pranita Menavlikar and Pavan Turlapati for conducting preliminary investigations that informed this work, and Logan Crowl for assisting with the implementation of the crowdsourced experiments, along with the human annotators who took part in the study. We also thank Dr. Naveen Sharma and Dr. Christian Newman for their support and feedback as part of Jonathan Pofcher's thesis committee. KhudaBukhsh was partly supported by a gift from Lenovo.

\clearpage


\bibliography{custom}
\clearpage
\appendix
\setcounter{table}{0}
\renewcommand{\thetable}{A\arabic{table}}

\section{Curating $\mathcal{D}_\textit{hope}$}

\subsection{Initial Human and LLM Annotation}\label{sec:human-LLM}
Initially, one of the authors label a set of 1,500 comments (500 randomly sampled comments from three news outlets hosting $\mathcal{V}_\textit{LGBTQ+}$ videos) as being hope speech directed towards the LGBTQ+ community or not. See Appendix A for the guidelines used in this process. This step only identified 80 positives (5.33\%) i.e., LGBTQ+ hope speech suggesting that hope speech is a rare class. In order to curate a balanced dataset in a time- and cost-effective way, we turn to leveraging human-LLM collaborative annotation in the following way~\cite{wang2024human}. Using the initial labelled set of 1,500 comments, we fine-tune \texttt{Llama2-13B} (training details are in the Appendix). While this fine-tuned classifier was not highly accurate, it allowed us to identify hope speech more efficiently. We use this model to classify sets of 20,000 unseen comments from each channel. The same author as before went through the positives and verified any true positives until we obtain a total of 975 verified LBGTQ+ hope speech comments; 325 from each channel. These comments combined with a set of 975 random negatives labeled by the model, balanced for each channel, made up the 1,950 comment corpus that would be labelled by our crowd-sourced annotators as our seed set. 

We devote a separate subsection to describe our crowd-sourced annotation study design. In what follows, we outline our active learning steps to expand our seed set.

\subsection{Active Learning}\label{sec:ActiveLearning}
We observe that our initial fine-tuned models performed considerably worse on the minority classes of \textit{Negative} and \textit{Neutral}, while they had relatively high performance on the \textit{Positive} and \textit{Irrelevant} classes. As a remedial measure, we employ two different active learning sampling strategies. 
\par The first was minority certainty sampling. Prior literature shows that this method can be useful for reducing class imbalance \cite{palakodety2020voice,attenberg-dual-supervision}. We consider our best-performing model at this point (a fine-tuned \texttt{Llama}) and have it classify an unseen set of 50k randomly sampled comments from each channel. We took the 150 comments of the highest certainty with labels \textit{Negative} and \textit{Neutral}, for each channel, totaling 900 instances.  We used the same crowd-sourced annotation design described in Section \ref{sec:CrowdsourcedAnnotation} to have these labelled. This allowed us to help close the gap in imbalance for these two labels. Our final dataset at this step was 2,850 instances. Looking at the distribution of comments with consensus among the annotators, we had 680 Positives, 682 Negatives, 227 Neutrals, and 764 Irrelevants. We found that this improved our model's performance, especially with the Neutral class, and were interested in seeing if continue sampling could improve it more.

\par Next we tried margin sampling. Using the now highest performing model, Llama trained on the certainty sampled data, we labelled 50,000 more comments from each channel. We looked at the margin of difference between the probability of the chosen label and the next highest one. We wanted to collect samples where this margin was lowest for each combination of labels, giving us a balanced look at where the model was uncertain. In total, we collected 25 comments for the 12 combinations of labels for each channel, adding up to 900 samples. Once again, we used the same crowd-sourced annotation design described in Section \ref{sec:CrowdsourcedAnnotation} to have these labelled. This was our final dataset of 3,750 instances, which consisted of 824 Positive, 947 Negative, 314 Neutral, and 1020 Irrelevant consensus labels.

\section{Comparison to Off-The-Shelf Classifiers}

Table~\ref{tab:baseline_comparison} compares the performance of our \textit{hope speech} classifier against traditional baselines for sentiment classification, hate speech detection, and stance detection.

\begin{table}[htb]
\centering
\tiny
\setlength{\tabcolsep}{6pt}
\begin{tabular}{lccc}
\toprule
\textbf{Classifier} & \textbf{Macro F1} & \textbf{Pos. F1} & \textbf{Neg. F1} \\
\midrule
Traditional Sentiment (DistilBERT-sst2) & 0.578 & 0.402 & -- \\
Traditional Sentiment (Positive/Negative only) & 0.601 & 0.52 & 0.682 \\ \color{black}
Toxicity-based (Detoxify-toxic-bert)         & 0.503 & --     & 0.129 \\
Hate Speech (cardiffnlp hate-roBERTa)              & 0.605 & --     & 0.258 \\
GPT-4o (Classification Prompt)                & 0.666 & 0.862 & 0.590 \\
Our best model (Fine-tuned Llama)            & 0.716 & 0.866  & 0.684 \\
\bottomrule
\end{tabular}
\caption{Performance comparison for off-the-shelf classifiers on relevant tasks. For classifiers optimized to detect supportive content, the Positive F1 is reported on a binary formulation mapping hope speech (Positive) versus non-hope
speech (Negative); for toxicity and hate speech detectors the Negative F1 is shown on a binary formulation mapping hate speech (Negative) versus non-hate
speech (Positive, Neutral, Irrelevant). Blank cells indicate metrics that do not apply.}
\label{tab:baseline_comparison}
\end{table}

\newpage

\section{Additional Related Work}
\subsection{Additional Literature on Counter Speech}

There exists a large body of research done into the automatic detection of hate-speech online. This is mainly relevant in the case of social media platforms such as Facebook or YouTube, where such hateful content can be very harmful to both user's online and in-person experiences, especially those belonging to marginalized or minority groups\cite{muller-karsten}. By identifying this hate-speech, either before or after it is posted, it can be censored or removed and the user's that posted such content can be warned or banned. In extreme cases, entire social media platforms can be taken down by those hosting them\cite{agarwal2022deplatforming}. More recent research has shown that censorship and suppression of offenders is not always the best way to deal with such situations. These approaches can often be unsuccessful in limiting the spread of hateful ideas and may even backfire, both in the context of social media and in society as a whole \cite{benesch2014countering}.
\par Counter-speech has been proposed as a method of mitigating the effects of and preventing hate-speech, without the need to limit free speech through censorship. Counter-speech is a somewhat vague term and current literature defines it in a multitude of ways and acknowledges that there are many different types and strategies of counter-speech that can be found online, each achieving varying levels of success in different communities. It was found that empathy-based approaches could cause posters of xenophobic hate-speech to delete their comments and be less likely to post similar content in the near feature \cite{hangartner2021}. However, it has also been shown that strategies such as utilizing a 'Positive Tone' or affiliating oneself with the marginalized group are often not effective and may even garner replies by other users stating that such comments will not change the opinions of the hate speakers \cite{mathew2019thou}. Another negative aspect of relying on counter-speech to moderate online communities is that it requires the existence of hate-speech to be employed. By most working definitions of counter-speech, it must be in reply to a hateful comment. This limits its effectiveness as it can only be used as a response and not proactively. 

\subsection{Additional Literature on Hope Speech}
Addressing the drawbacks of counter-speech and the complexities of censorship, the emerging field of hope speech research presents a promising approach. It focuses on promoting positive, inclusive dialogues, aiming to transform and de-escalate online communities into environments of support and mutual respect. The term hope speech started to be used a few years ago by authors looking for user posted content that aimed to diffuse conflict, for example, in the context of the tensions between Indian and Pakistan \cite{palakodety2019hope}. Before this, similar work had been done, but used other terms such as "help-speech" to refer to the positive discourse \cite{palakodety2020voice}. 
\par Since then, there have been multiple shared tasks in hope speech detection with over 50 teams presenting their findings \cite{chakravarthi-etal-2022-overview-shared} \cite{chakravarthi-muralidaran-2021-findings}. These shared tasks utilized datasets specifically created with hope-speech detection in mind, such as HopeEDI: a multi-lingual collection of almost 60,000 YouTube comments manually labeled as containing hope-speech or not \cite{chakravarthi-2020-hopeedi}. This dataset makes a point to contain hope speech directed at LGBTQ+ communities, among many other topics. Our study is different as it focuses directly on the LGBTQ+ community, but also views it through the lens of the US political divide. 
\par One related study that didn't explicitly focus on hope speech but provided valuable advancement to the field is the study of dehumanization. Dehumanization involves perceiving or treating people as less than human and often leads to extreme intergroup bias and hate speech. Mendelsohn, Tsvetkov, and Jurafsky developed a computational linguistic framework for analyzing dehumanizing language and applied it to discussions of LGBTQ people in the New York Times from 1986 to 2015 \cite{Mendelsohn_2020}. They found increasingly humanizing descriptions of LGBTQ people over time. Notably, different words with similar meanings, such as "gay" and "homosexual," had different levels of dehumanization, with "homosexual" being more associated with dehumanizing attitudes.
\par While much of the previous literature on hope speech detection mentions the LGBTQ+, as the task gravitates towards marginalized groups, there are gaps in research focusing directly on it. García‑Baena et al. created the dataset SpanishHopeEDI. It contains 1,650 annotated Spanish LGBTQ+-related tweets. A multitude of traditional machine-learning algorithms were tested on identifying hope-speech on them, leading to very impressive results \cite{garcia2023hope}. One related shared task similar to the previous ones was conducted focusing on this topic, however it was for hate-speech detection instead. For this, a dataset that specifically labelled LGBTQ+-related comments as homophobic or transphobic was used \cite{chakravarthi-etal-2022-overview}. 

\par Another important question for hope speech detection is understanding how subjective interpretations of language can impact the effectiveness of content moderation. A related study investigates the disagreement between human and machine moderators on what constitutes offensive speech in political discourse \cite{Vicarious}. This research highlights the challenges of subjectivity in moderation, revealing that both human and machine moderators often disagree on what is considered offensive, especially when political leanings are involved. These findings are relevant for hope speech detection as they underscore the need for nuanced understanding and handling of subjective content, ensuring that genuine hope speech is accurately identified amidst varying interpretations. In the same way that there can be disagreement on what is offensive, there can be disagreement on what is supportive or hopeful.

The recent introduction and consistent improvement of LLMs in the field of natural language processing has opened many new doors, and the task of hope speech detection is no different. However, at this point in time there has not been a lot of research into how effective LLMs are at hope-speech detection, never mind LGBTQ+ hope speech superficially. At the time of writing, only one paper has looked into this and attempted zero-shot ChatGPT prompting to label a Hope Speech dataset \cite{Ngo2023ZootopiAH}. They found that it resulted in poor performance for English texts, but was very successful for Spanish text. More work has been done surrounding LLM performance into hate-speech detection. The first extensive study into this found that prompting strategies had a significant impact on results \cite{guo2024investigationlargelanguagemodels}. They found using a Chain-of-Thought Reasoning Prompt had the highest results in all metrics and overall was effective in detecting hate-speech.


\newpage
\section{Prompts Used}\label{sec:Prompt}

\begin{figure}[H]
    \centering
    \includegraphics[width=\columnwidth]{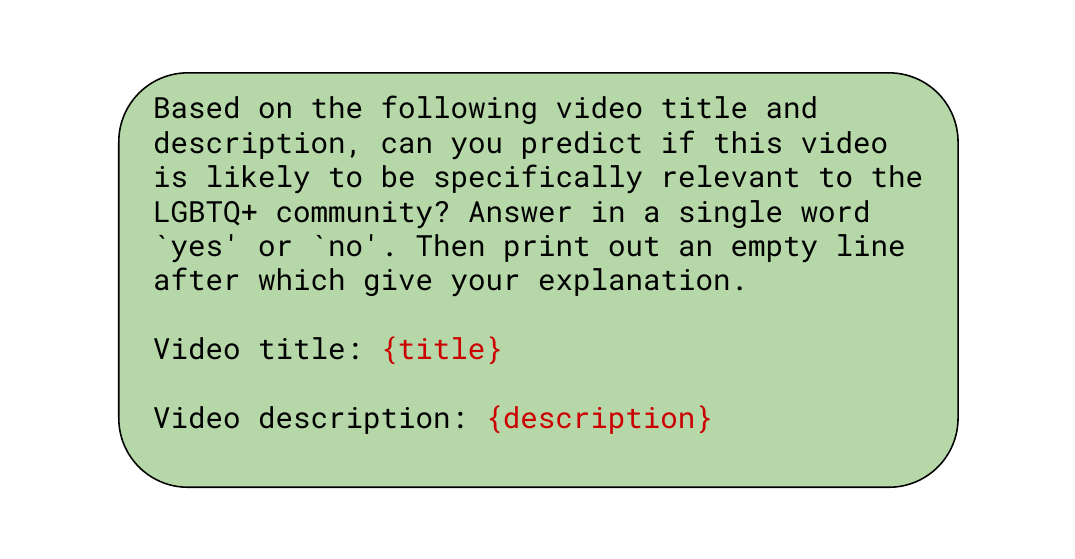}
    \caption{Prompt for LGBTQ+ Video Classification}
    \label{fig:lgbtq_prompt}
\end{figure}

\begin{figure}[H]
    \centering
    \includegraphics[width=\columnwidth]{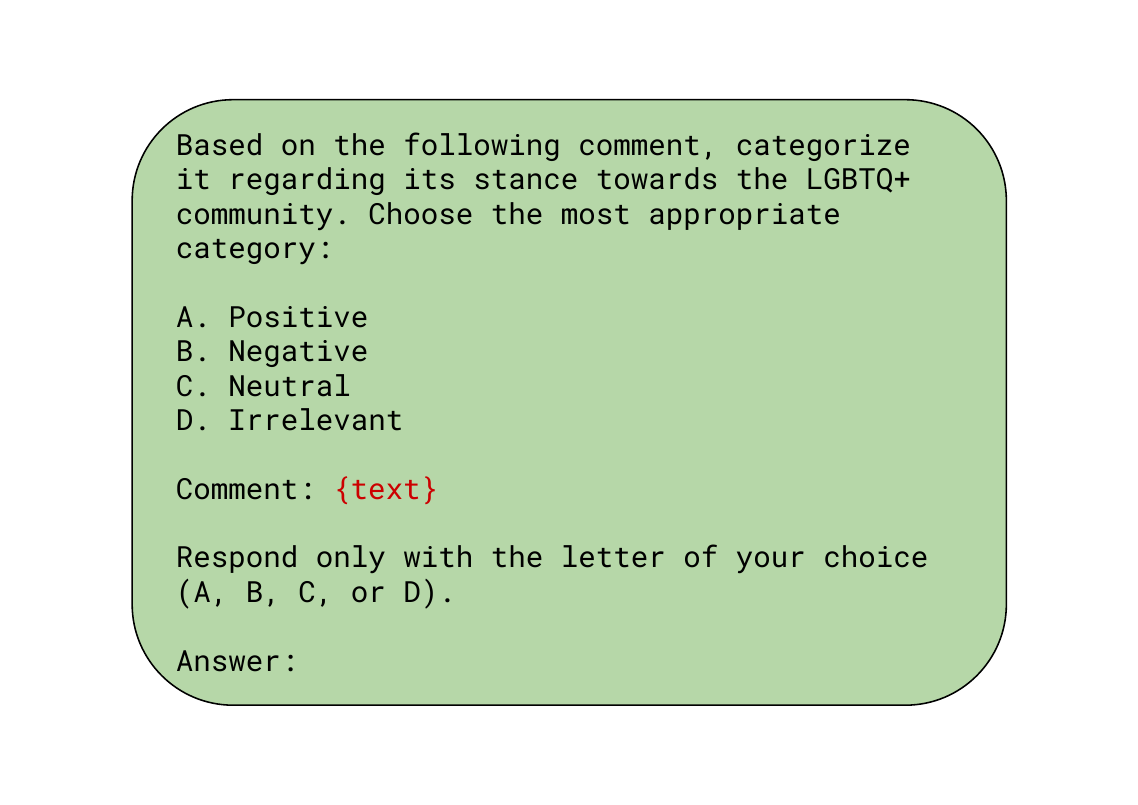}
    \caption{Prompt used for  classification}
    \label{fig:4label_prompt}
\end{figure}

\section{Annotation Guidelines}
\label{sec:guidelines}
\par A comment is marked as \textbf{Hope Speech} if it meets any of the following criteria:
\begin{enumerate}
    \item Advocates for LGBTQ+ well-being, rights, or acceptance. 
    \textit{E.g.,} "How can we support LGBTQ+ youth in America?"
    
    \item Urges support for LGBTQ+ rights or anti-discrimination efforts. 
    \textit{E.g.,} "Politicians need to be against the Don't Say Gay Bill."
    
    \item Pushes for LGBTQ+ equal rights (marriage, anti-discrimination, gender recognition). 
    \textit{E.g.,} "Trans rights are human rights, and it's time for our laws to reflect that."
    
    \item Denounces anti-LGBTQ+ violence, discrimination, or hate speech. 
    \textit{E.g.,} "Homophobes need to be stopped."
    
    \item Shows sympathy for LGBTQ+ struggles and solidarity. 
    \textit{E.g.,} "I'm straight, but love is love!"
    
    \item Indirectly supportive and inclusive of LGBTQ+ community. 
    \textit{E.g.,} "Everyone deserves to love who they love, without fear or judgment."
\end{enumerate}

\par A comment is marked as \textbf{Non-Hope Speech} if it meets any of the following criteria:
\begin{enumerate}
    \item Expresses violent intent or supports discriminatory practices against LGBTQ+. 
    \textit{E.g.,} "If my kid has a gay teacher, they better watch out."
    
    \item Calls for actions/policies harmful to LGBTQ+ rights or well-being. 
    \textit{E.g.,} "Marriage should only be between a man and a woman."
    
    \item Diverts from LGBTQ+ issues to unrelated topics, diminishing their importance. 
    \textit{E.g.,} "I'm all for same-sex marriage, but should we worry about world hunger before this stuff?"
    
    \item Spreads misinformation or stereotypes about LGBTQ+ community. 
    \textit{E.g.,} "The alphabet mob is brainwashing our kids!"
    
    \item Demonstrates sarcastic or mocking support for LGBTQ+. 
    \textit{E.g.,} "Everyone should be able to identify as anything they want. I identify as an attack helicopter!"
    
    \item Unrelated to LGBTQ+ issues. 
    \textit{E.g.,} "Abortion is a sin!"
\end{enumerate}

\section{Two-Label Agreement}
\begin{table}[h]
\centering
\scriptsize
\setlength{\extrarowheight}{2pt}
\begin{tabular}{cc|c|c|c|}
  & \multicolumn{1}{c}{} & \multicolumn{1}{c}{\textit{Dem}}  & \multicolumn{1}{c}{\textit{Rep}}  & \multicolumn{1}{c}{\textit{Ind}} \\\cline{3-5}
            & \textit{Dem} &\cellcolor{blue!25} - & 0.447  & 0.556
 \\ \cline{3-5}
 & \textit{Rep} & \textcolor{black}{0.447} &\cellcolor{blue!25} - & \textcolor{black}{0.432}
 \\\cline{3-5}
            & \textit{Ind} & \textcolor{black}{0.556} & \textcolor{black}{0.432}  &\cellcolor{blue!25} - \\\cline{3-5}
\end{tabular}
\caption{Two-Label Political Affiliation Agreement (Human Annotators) Values are Fleiss' $\kappa$}
\label{tab:politicalAgreement}
\end{table}

\begin{table}[h]
\centering
\scriptsize
\setlength{\extrarowheight}{2pt}
\begin{tabular}{cc|c|}
  & \multicolumn{1}{c}{} & \multicolumn{1}{c}{\textit{Non-LGBT}} \\\cline{3-3}
            & \textit{LGBT} & 0.498
 \\\cline{3-3}
\end{tabular}
\caption{Two-Label LGBTQ+ Agreement (Human Annotators) Values are Fleiss' $\kappa$}
\label{tab:lgbtAgreement}
\end{table}

\section{Two-Label Classification Prompt}

\begin{figure}[htb]
    \centering
    \includegraphics[width=\columnwidth]{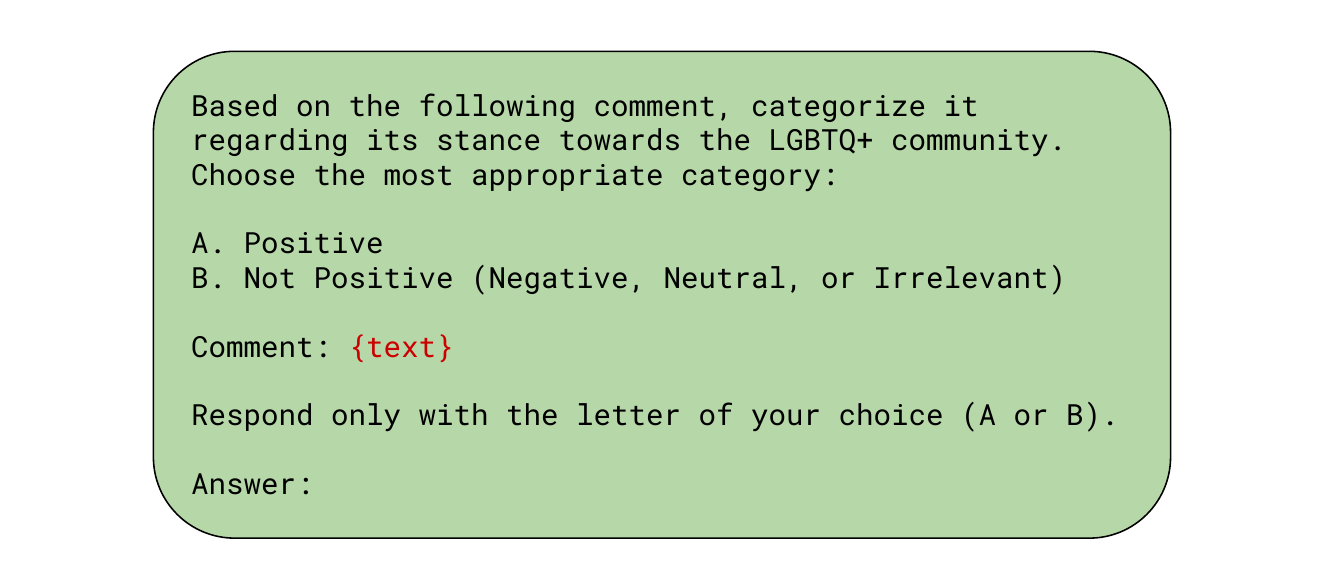}
    \caption{Prompt for 2-Label Classification}
    \label{fig:2label_prompt}
\end{figure}

The two-label classification prompt is presented in Figure~\ref{fig:2label_prompt}. 

\section{Crowd-sourced Study Compensation}
We compensate raters \$0.10 for each instance. This yields a compensation of \$3 for each 30 comment batch. We observe a median task completion time of 11:53 minutes, implying an hourly wage estimate of \$15.16, more than the minimum wage. Rater compensation is grounded in prior literature \cite{leonardelli-etal-2021-agreeing,bai-etal-2021-pre,Vicarious}. No participants complained about the compensation in their feedback.

\section{Survey Feedback}\label{homophobia}

Table \ref{tab:negative_feedback} lists a few examples of homophobic survey feedback. Table \ref{tab:positive_feedback} lists a few examples of survey feedback expressing a positive stance towards our research task or the LGBTQ+ community. 

\begin{table}[t]
\centering
\small
\begin{tabular}{|p{0.9\columnwidth}|}
\hline
\textbf{Examples of Negative Survey Feedback} \\
\hline
``You can identify as anything you want However how you were born cannot be changed or altered. Surgery and clothing does not change what you were born as. Time for common sense to reign again. 1,000 years from now when somebody digs up bones you will never hear them say oh look we just found a tr\censor{a}nny!!!!'' \\
\hline
``Comments speaking truth about the alphabet mafia are not negative just because the alphabet mafia does not like being confronted with truth.'' \\
\hline
\end{tabular}
\caption{Examples of survey feedback expressing homophobic or negative views.}
\label{tab:negative_feedback}
\end{table}

\begin{table}[t]
\centering
\small
\begin{tabular}{|p{0.9\columnwidth}|}
\hline
\textbf{Examples of Positive Survey Feedback} \\
\hline
``Reading so much hateful, ignorant rhetoric was tough! Thanks for all you're doing to spread peace and love in the world.'' \\
\hline
``This study was very eye opening. Thank you for letting me participate in this wonderful study. Have a wonderful day.'' \\
\hline
\end{tabular}
\caption{Examples of survey feedback expressing support for the research and LGBTQ+ community.}
\label{tab:positive_feedback}
\end{table}

\section{Annotator Demographics}

\begin{table}[h]
\centering
\begin{tabular}{|l|r|r|}
\hline
\textbf{Category} & \textbf{Count} & \textbf{Percentage} \\
\hline
\multicolumn{3}{|l|}{\textbf{Political Affiliation}} \\
\hline
Independent & 125 & 33.33\% \\
Democrat & 125 & 33.33\% \\
Republican & 125 & 33.33\% \\
\hline
\multicolumn{3}{|l|}{\textbf{LGBTQ+ Community Identity}} \\
\hline
Yes & 98 & 26.13\% \\
No & 277 & 73.87\% \\
\hline
\multicolumn{3}{|l|}{\textbf{Age Range}} \\
\hline
18-24 & 45 & 12.00\% \\
25-34 & 103 & 27.47\% \\
35-44 & 104 & 27.73\% \\
45-54 & 74 & 19.73\% \\
55-64 & 34 & 9.07\% \\
65 or older & 15 & 4.00\% \\
\hline
\multicolumn{3}{|l|}{\textbf{Self Description}} \\
\hline
Male & 153 & 40.80\% \\
Female & 209 & 55.73\% \\
Nonbinary/third gender & 10 & 2.67\% \\
Self-describe & 1 & 0.27\% \\
Prefer not to say & 2 & 0.53\% \\
\hline
\end{tabular}

\caption{Demographic Breakdown of Survey Respondents (Total Responses: 375)}
\label{tab:demographic-breakdown}
\end{table}
All annotators were living in the United States. Complete demographics and a breakdown of LGBTQ+ demographics by political affiliation can be found in Table \ref{tab:demographic-breakdown} and \ref{tab:political-lgbt-breakdown}

\FloatBarrier

\begin{table}[h]

\centering
\begin{tabular}{|>{\color{black}}l|>{\color{black}}r|>{\color{black}}r|}
\hline
{\textbf{Political}} & {\textbf{LGBTQ+}} & {\textbf{Not LGBTQ+}} \\
\textcolor{black}{\textbf{Affiliation}} & \textcolor{black}{\textbf{Count (\%)}} & \textcolor{black}{\textbf{Count (\%)}} \\
\hline
\textcolor{black}{Independent} & \textcolor{black}{38 (30.4\%)} & \textcolor{black}{87 (69.6\%)} \\
\textcolor{black}{Democrat} & \textcolor{black}{31 (24.8\%)} & \textcolor{black}{94 (75.2\%)} \\
\textcolor{black}{Republican} & \textcolor{black}{29 (23.2\%)} & \textcolor{black}{96 (76.8\%)} \\
\hline
\textcolor{black}{\textbf{Total}} & \textcolor{black}{98 (26.1\%)} & \textcolor{black}{277 (73.9\%)} \\
\hline
\end{tabular}
\caption{Distribution of LGBTQ+ Identity Across Political Affiliations}
\label{tab:political-lgbt-breakdown}
\end{table}

\section{Comparison Across Controversial Topics}\label{sec:controversial-topics}
\begin{table*}[h]
\centering
\small
{%
  \begin{tabular}{|l|cc|cc|cc|}
  \hline
  & \multicolumn{2}{c|}{\textbf{MSNBC}} & \multicolumn{2}{c|}{\textbf{FOX}} & \multicolumn{2}{c|}{\textbf{CNN}} \\
  \textbf{Topic} & \textbf{Mean} & \textbf{Med.} & \textbf{Mean} & \textbf{Med.} & \textbf{Mean} & \textbf{Med.} \\
  \hline
  Control & 21.5\% & 13.5\% & 19.4\% & 10.9\% & 27.1\% & 20.3\% \\
  LGBTQ+ & 24.1\% & 18.8\% & 29.7\% & 20.8\% & 35.5\% & 32.4\% \\
  Gun & 24.1\% & 16.3\% & 22.8\% & 15.5\% & 37.9\% & 33.3\% \\
  Climate & 17.0\% & 10.6\% & 21.7\% & 11.4\% & 34.6\% & 32.0\% \\
  Abortion & 20.3\% & 12.4\% & 17.6\% & 8.0\% & 29.0\% & 23.4\% \\
  Immigration & 19.5\% & 14.2\% & 16.9\% & 8.0\% & 40.8\% & 38.7\% \\
  \hline
  \end{tabular}%
}
\caption{Dislike ratio comparison of YouTube videos across controversial topics by news outlet. For each topic, we show the mean and median (Med.) dislike ratios.}
\label{tab:controversial-topics}
\end{table*}

In an attempt to expand upon the findings in Table \ref{tab:dislike-engagement-rates}, we analyze dislike ratios across multiple hot-button issues in Table \ref{tab:controversial-topics}. While many of  the topics elicit more polarized engagement than control content, LGBTQ+ content shows distinct patterns: it ties for the most polarizing content on MSNBC (24.1\% mean dislike ratio), is considerably more polarizing on Fox News (29.7\%, significantly exceeding traditional controversial topics like gun control at 22.8\% and climate change at 21.7\%), and maintains the third-highest negative engagement on CNN (35.5\%) after immigration (40.8\%) and gun-related content (37.9\%). These findings demonstrate that LGBTQ+ topics generate uniquely high negative engagement across all platforms, regardless of the channel's political leaning. Although this is more pronounced on FOX, and less pronounced on CNN, which receives high dislike rates compared to the other channels across all topics.

\section{Analysis of Video Content by Positivity Ratio}
\label{sec:video-content}
\begin{table*}[t]
\centering
\footnotesize
\begin{tabular}{|l|l|p{11.5cm}|}
\hline
\textbf{Channel} & \textbf{Category} & \textbf{Representative Videos} \\
\hline
\multirow{2}{*}{CNN} & Most Positive & 
1. Biden's surprise brings Elton John to tears at White House \\
& & 2. Pete Buttigieg reacts to Martha Alito's Pride flag comment \\
& & 3. Supreme Court says federal law protects LGBTQ workers from discrimination \\
& & 4. Church member defends Pastor Worley's Anti-Gay Rant \\
& & 5. GOP Rep. attends gay son's wedding after opposing protections \\
\cline{2-3}
& Least Positive & 
1. Chicago police slam decision in Jussie Smollett case \\
& & 2. Zoey Tur on Transgender Rights \\
& & 3. Watch the moment Brittney Griner lands on US soil \\
& & 4. FILE: JOAN RIVERS JOKES MRS OBAMA "IS A TRANS" \\
& & 5. See confrontation between Ben Carson and lawmaker over transgender rights \\
\hline
\multirow{2}{*}{FOX} & Most Positive & 
1. Santorum answers gay soldier's DADT question \\
& & 2. Franklin Graham: Christians should prepare for persecution after gay marriage ruling \\
& & 3. Huckabee: 'This is where we've gone to la la land' \\
& & 4. Trey Gowdy, Kayleigh McEnany rip liberals for attacking Caitlyn Jenner \\
& & 5. 'OUTRAGE': Biden honors 'Trans Day of Visibility' on Easter Sunday \\
\cline{2-3}
& Least Positive & 
1. Democratic Socialist convention erupts over pronouns \\
& & 2. Gutfeld: How did this nutcase get a security clearance? \\
& & 3. 'Turned into a circus': Democrats storm out of gender reassignment hearing \\
& & 4. Target insider sounds alarm: 'Terrified' of Bud Light-style backlash \\
& & 5. Gutfeld: Biden official accused of second luggage heist \\
\hline
\multirow{2}{*}{MSNBC} & Most Positive & 
1. Internet Sensation Randy Rainbow On Florida's 'Don't Say Gay' Bill \\
& & 2. Maddow: We Feared Susan's Covid Would Kill Her \\
& & 3. Federal Judge Rules Anti-HIV Medicine Is Unconstitutional \\
& & 4. Drag queens describe a Pride Month like no other \\
& & 5. See GOP confronted over 'straights only' discrimination \\
\cline{2-3}
& Least Positive & 
1. Jen Psaki Leaves White House \\
& & 2. Jotaka Eaddy: Criticizing Brittney Griner's Return Is 'Un-American' \\
& & 3. John Heilemann: What's the body count on Critical Race Theory? \\
& & 4. Brittney Griner Freed As Part Of Russian Prisoner Swap \\
& & 5. For Facts Sake: Florida's Parental Rights In Education Bill \\
\hline
\end{tabular}
\caption{Most and least positive LGBTQ+ videos by channel, ranked by user engagement positivity ratio. Videos selected from those with at least 100 comments and at least 10 positive/negative comments.}
\label{tab:video-content}
\end{table*}
In order to examine the tone and content of videos that elicit highly positive or negative responses towards the LGBTQ+ community, we find the videos with the highest and lowest positivity ratio for each channel in Table \ref{tab:video-content}
Our analysis shows distinct patterns in how different channels cover LGBTQ+ topics. Fox News videos maintain an opposing stance regardless of positivity ratio, with only one positive video appearing to feature a balanced perspective. CNN and MSNBC's most positive content focuses on pop culture and celebrity figures (e.g., Elton John, Randy Rainbow, Maddow). Gender identity and transgender issues appear disproportionately in lowest-positivity content across channels (3/5 CNN, 5/5 FOX, 1/5 MSNBC videos). One interesting finding is that a very specific news story, coverage of Brittney Griner's release, consistently generates lower positivity ratios, appearing in 3 of the least positive videos.

\section{Lexical Analysis of Comments by Category}
\label{sec:lexical-analysis}

In order to examine the type of discussion happening in these comments, we performed a lexical analysis in different comment categories based on aggregate crowd-sourced labels. For each category comparison (e.g., positive vs. negative), we computed log odds ratios with Laplace smoothing to identify words that appear disproportionately in one category versus another. A positive log odds ratio indicates the word is more characteristic of the first category, while a negative ratio indicates association with the second category. The magnitude of the ratio reflects the strength of this association. This analysis reveals patterns in how language is utilized across supportive, critical, neutral, and off-topic comments.

\subsection{Positive vs. Negative Comments}
Table~\ref{tab:pos_vs_neg_full} presents the distinctive words for positive and negative comments. On the positive side, one of the most clear patterns is the presence of terms such as \textit{constitution}, \textit{freedoms}, \textit{citizen}, \textit{fundamental}, and \textit{defending}. These words suggest that supportive comments often reference the protection of human rights and civil liberties in the defense of the LGBTQ+. 

In contrast, the negative comments are most clearly distinguished by words like \textit{holy}, \textit{sodom}, \textit{gomorrah}, \textit{leviticus}, and \textit{repent}. This pattern suggests that opposition to LGBTQ+ rights often stems from religious frameworks and traditional moral values, sometimes even directly citing the Bible. The stark difference in these foundational frameworks—legal rights versus religious doctrine—highlights a fundamental tension in how different groups conceptualize LGBTQ+ issues.

One additional divergence in vocabulary between these two group may be rather obvious, but can be seen in the tone of words used. Positive comments frequently employ terms of inclusivity and understanding such as \textit{love}, \textit{wonderful}, \textit{cared}, \textit{understanding}. Negative comments are marked by language of moral condemnation such as \textit{perversion}, \textit{insanity}, \textit{immorality}, and \textit{unfair}. This shows how these categories are more than just "stance on LGBTQ+" and how one side is dehumanizing the LGBTQ+ community, while the other is humanizing it.

Finally, the political dimension of this discourse is evident in the distinctive use of political identifiers. The presence of \textit{rightwing} in positive comments, contrasted with \textit{democrat}) and \textit{democrats} in negative ones, suggests that supporters of LGBTQ+ rights often frame their opposition in terms of right-wing ideology, while those expressing negative sentiments frequently attribute their grievances to Democratic politics.

\subsection{Irrelevant vs. On-Topic Comments}
In Table~\ref{tab:irr_vs_topic_full}, we compare vocabulary between irrelevant comments and those that are on-topic. As with neutral comments, it is somewhat difficult to identify a pattern in the distinctively irrelevant words, as it encompasses such a large spectrum.  Irrelevant comments often contain references to other politicized topics, with terms like  \textit{newsom}, \textit{covid}, \textit{blm}, and \textit{migrants}.  This pattern suggests that discussions of LGBTQ+ issues often become entangled with broader political debates and culture war topics.

On-topic comments, conversely, maintain focus on LGBTQ+-specific terminology and issues, with words almost all words being related such as  \textit{lgbt}, \textit{sexuality}, and \textit{transgender}.
This clear lexical differentiation helps validate our classifications and highlights how political discourse can drift across topics.

\begin{table}[t]
\centering
\footnotesize
\begin{tabular}{@{}ll@{}}
\toprule
\multicolumn{2}{c}{\textbf{Distinctive Words for Positive vs. Negative Comments}} \\ \midrule
\textbf{Top 30 words (Positive)} & \textbf{Top 30 words (Negative)} \\ \midrule
constitution (2.254)   & looks (–3.035) \\
species (2.053)        & perversion (–2.709) \\
spread (2.053)         & mankind (–2.542) \\
scary (2.053)          & company (–2.447) \\
affect (2.053)         & democrat (–2.342) \\
condition (2.053)      & letting (–2.342) \\
fully (2.053)          & holy (–2.342) \\
love (2.018)           & cake (–2.342) \\
roman (1.935)          & flesh (–2.284) \\
rightwing (1.935)      & logic (–2.224) \\
persecute (1.935)      & insanity (–2.224) \\
hypocrisy (1.935)      & west (–2.224) \\
wonderful (1.935)      & services (–2.224) \\
neighbors (1.935)      & sodom (–2.224) \\
treat (1.871)          & leviticus (–2.224) \\
experiences (1.802)    & aint (–2.159) \\
cared (1.802)          & masculine (–2.090) \\
akon (1.802)           & professor (–2.090) \\
activity (1.802)       & sacred (–2.090) \\
valid (1.802)          & gomorrah (–2.090) \\
citizen (1.802)        & muslim (–2.090) \\
persecuted (1.802)     & space (–2.090) \\
freedoms (1.802)       & unfair (–2.090) \\
understanding (1.802)  & gain (–2.090) \\
existence (1.802)      & lust (–2.090) \\
treated (1.765)        & libraries (–2.090) \\
beings (1.727)         & financial (–2.090) \\
defending (1.647)      & repent (–2.090) \\
fundamental (1.647)    & democrats (–2.090) \\
testicles (1.647)      & immorality (–2.090) \\ \bottomrule
\end{tabular}
\caption{Distinctive words for Positive versus Negative labeled comments. Numbers in parentheses are the log odds ratios.}
\label{tab:pos_vs_neg_full}
\end{table}

\begin{table}[t]
\centering
\footnotesize
\begin{tabular}{@{}ll@{}}
\toprule
\multicolumn{2}{c}{\textbf{Distinctive Words for Irrelevant vs. On-Topic Comments}} \\ \midrule
\textbf{Top 30 words (Irrelevant)} & \textbf{Top 30 words (On-Topic)} \\ \midrule
yahuah (4.513)                & transgender (–3.014) \\
yahushua (3.889)              & lgbt (–2.913) \\
yahuahs (3.579)              & gays (–2.799) \\
xxxxxxxxxxxxxxxxxxxxxxxxx (2.973) & homosexual (–2.770) \\
survey (2.973)               & gay (–2.726) \\
newsom (2.973)               & lgbtq (–2.716) \\
cookies (2.790)              & compete (–2.470) \\
hoping (2.790)               & heterosexual (–2.345) \\
covid (2.790)                & homophobia (–2.285) \\
misty (2.790)                & sexuality (–2.237) \\
obedience (2.790)            & lesbian (–1.997) \\
brittany (2.790)             & bigoted (–1.864) \\
susan (2.790)                & gender (–1.774) \\
commandment (2.790)          & shooter (–1.764) \\
teenagers (2.790)            & dysphoria (–1.764) \\
gavin (2.790)                & surgery (–1.764) \\
rapists (2.790)              & visibility (–1.764) \\
slaves (2.685)               & trans (–1.745) \\
rico (2.567)                 & transgenders (–1.737) \\
disapproval (2.567)          & sports (–1.709) \\
material (2.567)             & flesh (–1.709) \\
biracial (2.567)             & community (–1.681) \\
sources (2.567)              & celebrate (–1.652) \\
thanksgiving (2.567)         & students (–1.652) \\
gift (2.567)                 & basic (–1.652) \\
blm (2.567)                  & cis (–1.652) \\
60s (2.567)                  & consider (–1.652) \\
migrants (2.567)             & womens (–1.640) \\
139 (2.567)                  & weird (–1.592) \\
mob (2.567)                  & transphobic (–1.592) \\ \bottomrule
\end{tabular}
\caption{Distinctive words for Irrelevant versus on-topic (Positive+Negative+Neutral) comments. Numbers in parentheses are the log odds ratios.}
\label{tab:irr_vs_topic_full}
\end{table}

\clearpage
\section{Error Analysis}
\label{error-analysis}

While our model performed well in identifying the rare class of LGBTQ+ hope speech in the wild, there were cases where the beliefs expressed in comments were too nuanced to be classified correctly. Table \ref{tab:error-examples} showcases two such examples. Both comments are generally supportive on a surface level. One is promoting human rights and social progress, the other is giving respect to female athletes. However, in context it is clear the support is not directed to the LGBTQ+. The first comment is equating the social progress of the LGBTQ+ community with bestiality, a common homophobic cliché. The second comment is specifically singling out "biological female athletes" to support, taking a clear stance on the issue of transgender people in sports.
\par Examples like these illustrate the challenges a task like this entails. Our model fails to fully understand the sarcasm or superficial support that less straight-forward comments may contain. Enhancing the model's ability to understand context and implicit meanings, possibly through the use of more advanced language models or improved tuning  merits deeper exploration. 

\begin{table}[t]
\centering
\small
\begin{tabular}{|p{0.25\textwidth}|p{0.12\textwidth}|}
\hline
\textbf{YouTube Comment} & \textbf{Model Prediction} \\
\hline
\textit{Bestiality rights are human rights. Vote biden to continue social progress.} & Positive \\
\hline
\textit{Power and respect to her and all other biologically female athletes. STAY LOUD} & Positive \\
\hline
\end{tabular}
\caption{\small{Two example challenging comments our classifier mistakenly identified as Positive.}}
\label{tab:error-examples}
\end{table}

\begin{table}[t]
\centering
\begin{tabular}{|p{0.25\textwidth}|p{0.08\textwidth}|p{0.08\textwidth}|}
\hline
\textbf{Comment} & \textbf{LGBT} & \textbf{Non-LGBT} \\
\hline
\textit{Imagine how much faster HE would be in the water, if he didn't have his junk slowing him down!} & Negative & Irrelevant \\
\hline
\end{tabular}
\caption{\small{Example human annotations showing differences between LGBT and non-LGBT annotators.}}
\label{tab:human-examples}
\end{table}

\begin{table}[t]
\centering
\begin{tabular}{|p{0.15\textwidth}|p{0.10\textwidth}|p{0.12\textwidth}|}
\hline
\textbf{Comment} & $\mathcal{M}_\textit{LGBTQ+}$ & $\mathcal{M}_\textit{non-LGBTQ+}$ \\
\hline
\textit{Now why did God destroy Sodom and Gomorrah? Hmmmmm.} & Negative & Irrelevant \\
\hline
\end{tabular}
\caption{\small{Example predictions from models trained on LGBTQ+ and non-LGBTQ+ annotated data (denoted by $\mathcal{M}_\textit{LGBTQ+}$ and $\mathcal{M}_\textit{non-LGBTQ+}$, respectively ).}}
\label{tab:model-examples}
\end{table}

\section{Licenses}
Meta Llama 3 is used under the Meta Llama 3 Community License. Mistral is used under the Mistral AI Non-Production License.

\section{Computational Resources}
Models were trained using both 7B (Mistral) and 8B (Llama) parameter versions. Training was done  on a university-ran high-performance computing cluster using a single A100 GPU. Experiments totalled approximately 120 GPU hours.

\section{Packages}
\begin{table}[h]
\centering
\begin{tabular}{|l|l|}
\hline
\textbf{Package} & \textbf{Version} \\
\hline
PyTorch & 2.3.0-rc12 \\
Transformers & 4.35.2 \\
Datasets & 2.8.0 \\
NumPy & 1.26.3 \\
Scikit-learn & 1.4.0 \\
PEFT & 0.5.0 \\
\hline
\end{tabular}
\caption{Software packages and versions}
\label{tab:package-versions-simple}
\end{table}

\section{Training Setup and Hyperparameters}
\begin{table}[h]
\centering
\begin{tabular}{|l|l|}
\hline
\textbf{Hyperparameter} & \textbf{Value} \\
\hline
Learning Rate & 2e-4 \\
Batch Size & 8 \\
Number of Epochs & 5 \\
Weight Decay & 0.01 \\
LoRA rank (r) & 8 \\
LoRA alpha & 32 \\
LoRA dropout & 0.1 \\
\hline
\end{tabular}
\caption{Hyperparameters used during our fine-tuning}
\label{tab:hyperparameters}
\end{table}

We used the PEFT library for parameter-efficient fine-tuning with LoRA. Model selection was done using the best macro F1 score on the evaluation dataset.

\clearpage
\section{LGBTQ+ Label Distribution Comparison}

Table~\ref{tab:lgbtq-analysis} contrasts the label distributions obtained from LGBTQ+ and non-LGBTQ+ raters across different political affiliations. 

\begin{table*}
\centering
\small
\begin{tabular}{|l|l|c|c|r|c|}
\hline
\textbf{Political Affiliation} & \textbf{Label Type} & \textbf{LGBTQ+} & \textbf{Non-LGBTQ+} & \textbf{Difference} & \textbf{\textit{p}-value} \\
\hline
\multirow{4}{*}{Democrat} & Positive & 25.3\% & 24.3\% & +1.0\% & 0.543 \\
& Negative & 29.6\% & 29.1\% & +0.6\% & 0.753 \\
& Neutral & 17.7\% & 15.2\% & +2.5\% & 0.078 \\
& Irrelevant & 27.4\% & 31.5\% & -4.1\% & 0.023* \\
\hline
\multirow{4}{*}{Republican} & Positive & 21.7\% & 26.1\% & -4.3\% & 0.012* \\
& Negative & 36.8\% & 27.7\% & +9.1\% & <0.001*** \\
& Neutral & 16.5\% & 15.3\% & +1.2\% & 0.405 \\
& Irrelevant & 24.9\% & 30.8\% & -6.0\% & 0.001** \\
\hline
\multirow{4}{*}{Independent} & Positive & 27.7\% & 24.7\% & +3.0\% & 0.059 \\
& Negative & 30.1\% & 28.6\% & +1.5\% & 0.370 \\
& Neutral & 14.1\% & 15.1\% & -1.0\% & 0.438 \\
& Irrelevant & 28.1\% & 31.6\% & -3.5\% & 0.038* \\
\hline
\multicolumn{6}{l}{\small *\textit{p} < 0.05, **\textit{p} < 0.01, ***\textit{p} < 0.001}
\end{tabular}
\caption{Label distribution by political affiliation and LGBTQ+ identity. Values show the percentage of comments assigned each label type, with z-tests used to compute significance of differences.}
\label{tab:lgbtq-analysis}
\end{table*}

\section{LGBTQ+ Video Classification Pipeline Examples}

Table~\ref{tab:GPTExplanation} shows an example true positive and a false positive found in our results. The first shows a case in which the video ended up not being LGBTQ+ relevant, except for the fact that the person featured is a director of trans care, but the video does not focus on that fact. It is interesting that \texttt{GPT-4o} actually acknowledges this in its explanation, however did end up still flagging it as relevant. The second is an example that identifies the \textit{bathroom bill} as relevant to LGBTQ+ issues.

\begin{table*}[htb]
\scriptsize
\setlength{\tabcolsep}{2pt}
\renewcommand{\arraystretch}{1.2}
\begin{tabular}{|p{0.22\textwidth}|p{0.28\textwidth}|p{0.39\textwidth}|p{0.11\textwidth}|}
\hline
\textbf{Video title} & \textbf{Video description} & \textbf{\texttt{GPT-4o} explanation} & \textbf{Classification} \\
\hline
\rowcolor{gray!15}
Attacks On Abortion Rights Already Forcing Women To Dangerous 'Back Alley' Alternatives & 
Dr. Bhavik Kumar, medical director of primary and trans care at Planned Parenthood Gulf Coast, talks with Rachel Maddow about the desperation and hardship the Texas abortion ban is causing people seeking abortion services to which they have a right, and explains that the longer term restrictions on reproductive rights are already forcing people seeking abortions into dangerous circumstances. & 
The video features Dr. Bhavik Kumar, who is the medical director of primary and trans care at Planned Parenthood Gulf Coast. Although the main topic of the video is about the attacks on abortion rights in Texas, the mention of "trans care" in Dr. Kumar's professional role suggests that the topic might be relevant to transgender people, who are a part of the LGBTQ+ community. However, since the description doesn't explicitly state this, it's not a certainty, but a possibility. & 
False Positive \\
\hline
\rowcolor{white!15}
Montel Williams slams North Carolina's 'bathroom bill' &
Activist talks problematic articles within the controversial legislation on 'The Kelly File' &
The video seems to be relevant to the LGBTQ+ community based on the reference to North Carolina's 'bathroom bill,' a controversial legislation that affected the rights of transgender people, which makes it pertinent to their interests and concerns. Montel Williams being noted as an activist, alongside mentioning 'problematic articles' in the bill, suggests a discussion about these LGBTQ+ issues. &
True Positive \\
\hline
\end{tabular}
\caption{Videos identified as LGBTQ+ relevant by our \texttt{GPT-4} pipeline. The top example shows a case in which the video ended up not being LGBTQ+ relevant, except for the fact that the person featured is a director of trans care, but the video does not focus on that fact. It is interesting that \texttt{GPT-4o} actually acknowledges this in its explanation, however did end up still flagging it as relevant. The second is an example that identifies the \textit{bathroom bill} as relevant to LGBTQ+ issues.}
\label{tab:GPTExplanation}
\end{table*}

\section{In-the-wild Label Distribution}
\begin{table*}[t]
\centering
\footnotesize
\label{tab:label-breakdown-with-ci}
\begin{tabular}{lrrrr}
\hline
\textbf{Label} & \textbf{Overall} & \textbf{CNN} & \textbf{FOX} & \textbf{MSNBC} \\
\hline
Irrelevant & 76.53\% ± 0.025\% & 72.96\% ± 0.06\% & 71.78\% ± 0.045\% & 84.87\% ± 0.045\% \\
Negative   & 16.90\% ± 0.025\% & 17.20\% ± 0.05\% & 24.15\% ± 0.045\% & 9.36\% ± 0.04\% \\
Positive   & 4.84\% ± 0.015\% & 7.66\% ± 0.035\% & 2.30\% ± 0.02\% & 4.55\% ± 0.02\% \\
Neutral    & 1.73\% ± 0.01\% & 2.19\% ± 0.02\% & 1.77\% ± 0.02\% & 1.22\% ± 0.01\% \\
\hline
Positivity Ratio  & 22.25\% ± 0.065\% & 30.81\% ± 0.12\% & 8.70\% ± 0.07\% & 32.70\% ± 0.15\% \\
\hline
\end{tabular}
\caption{{Label breakdown and positivity rates by channel with 95\% confidence intervals. 50k comments from each channel
found in-the-wild were classified by our best performing model.}}
\label{tab:in-the-wild}
\end{table*}

\section{Subreddit Label Distribution}
\begin{table*}[t]
\centering
\footnotesize
\begin{tabular}{lrrrr}
\hline
\textbf{Label} & \textbf{Overall} & \textbf{r/democrat} & \textbf{r/republican} & \textbf{r/politics} \\
\hline
Irrelevant & 80.61\% & 85.87\% & 77.89\% & 78.06\% \\
Negative   & 9.01\% & 5.04\% & 14.52\% & 7.48\% \\
Positive   & 6.17\% & 5.88\% & 3.47\% & 9.15\% \\
Neutral    & 4.21\% & 3.21\% & 4.11\% & 5.31\% \\
\hline
Positivity Ratio  & 40.64\% & 53.83\% & 19.29\% & 55.01\% \\
\hline
\end{tabular}
\caption{ Label breakdown by subreddit. 15k comments from each subreddit were classified by our best performing model}
\label{tab:subreddit-distribution}
\end{table*}

\end{document}